\documentclass{article}

\PassOptionsToPackage{numbers, compress, square}{natbib}
 \usepackage[preprint]{neurips_2025}

\usepackage[utf8]{inputenc} 
\usepackage[T1]{fontenc}    
\usepackage{hyperref}       
\usepackage{url}            
\usepackage{booktabs}       
\usepackage{amsfonts}       
\usepackage{nicefrac}       
\usepackage{microtype}      
\usepackage{xcolor}         
\usepackage{graphicx}
\usepackage{array}
\usepackage{multirow}
\usepackage{nameref}
\usepackage{varioref}
\usepackage{hyperref}
\usepackage{cleveref}

\usepackage{colortbl}
\usepackage{booktabs}
\usepackage{caption}
\usepackage{soul}
\usepackage{subcaption}
\title{Examining Vision Language Models through Multi-dimensional Experiments with Vision and Text Features}

\author{%
Saurav Sengupta$^{1}$ \quad Nazanin Moradinasab$^{1}$ \quad Jiebei Liu$^1$ \quad Donald E. Brown$^1$ \\
$^1$School of Data Science, University of Virginia \\
\texttt{\{ss4yd,nm4wu,mcu2xn,deb\}@virginia.edu}\\
}

\begin{document}

\maketitle

\begin{abstract}
  Recent research on Vision Language Models (VLMs) suggests that they rely on inherent biases learned during training to respond to questions about visual properties of an image. These biases are exacerbated when VLMs are asked highly specific questions that require focusing on specific areas of the image. For example, a VLM tasked with counting stars on a modified American flag (e.g., with more than 50 stars) will often disregard the visual evidence and fail to answer accurately. We build upon this research and develop a multi-dimensional examination framework to systematically determine which characteristics of the input data, including both the image and the accompanying prompt, lead to such differences in performance. Using open-source VLMs, we further examine how attention values fluctuate with varying input parameters (e.g., image size, number of objects in the image, background color, prompt specificity). This research aims to learn how the behavior of vision language models changes and to explore methods for characterizing such changes. Our results suggest, among other things, that even minor modifications in image characteristics and prompt specificity can lead to large changes in how a VLM formulates its answer and, subsequently, its overall performance. 

\end{abstract}

\vspace{-0.5cm}
\section{Introduction and motivation}

\vspace{-0.3cm}
Recent research by Vo et al.\citep{vlmsarebiased} shows that state-of-the-art Vision Language Models (VLMs) like Open AI's o3 \cite{o3} and Google DeepMind's Gemini 2.5 Pro \cite{gemini} are biased due to memorization of prior knowledge. This leads to a severe decrease in accuracy in tasks that require precise counts. The authors explore this bias in real-world images with altered animal images where 1 extra leg is added to 2-legged or 4-legged animals, and images of well-known brand logos that are altered subtly. Prior knowledge learned through the massive text and image corpus these models are trained on predisposes them towards well-known answers rather than evaluating the image on its visual features, which in turn reduces counting accuracy.
Similar research has examined the vision understanding of VLMs by evaluating their Visual Question Answering (VQA) capabilities using synthetic imaging data \cite{hou2024vision, lee2024vlind, lee2024vhelm} or visualizing Attention Guided class activation maps (AG-CAM) on chart-based data \cite{dong2025probing}. Previous works have also examined the role of attention in VLM performance. VLMs have been shown to attend to certain visual tokens more than others \cite{woo2024don, an2025mitigating}. Kang et al. \cite{kang2025see} show that VLMs give more attention weight to irrelevant sections of the image and propose a method to redistribute the attention to more relevant areas.

While we replicate results from Vo et al. \cite{vlmsarebiased} in open source models like Qwen 2.5 VL \cite{qwen2.5-VL} and Kimi-VL-A3B \cite{kimiteam2025kimivltechnicalreport}, we also see that changing the prompt to focus on specific areas of the image can remove learned bias and help focus the VLM to answer the question correctly (Appendix Section \ref{sec:a1}). Examining this phenomenon in detail is the main motivation for this paper. Prior research is often focused on developing methods to evaluate VLM performance on an uni-dimensional variable, i.e. by varying image inputs and a set of associated factual questions with known answers that can be compared against VLM outputs. We believe there is a need to develop a multidimensional \textit{examination} framework that can be utilized to understand the effect of semantic variations in text and differences in visual characteristics of the image on the internal state and outputs of a VLM. We believe that this will lead to important insights as to how the model interprets input data under different scenarios and potentially lead to exploring areas of improvement. To that end we introduce a framework focusing on the multiple image and text based variables (e.g., prompt specificity, foreground and background color etc.) and examine their effect on VLM performance. We prioritize accuracy in object counting as a surrogate measure of a VLM’s response to the prompt. Additionally, we assess the distribution of attention to vision and text-based tokens to gain insights into how semantic alterations in text and visual transformations in images can result in varying model behaviors.

\vspace{-0.5cm}
\section{Methodology}

\vspace{-0.45cm}
\begin{figure}[htbp]
\centering
\includegraphics[width=\linewidth]{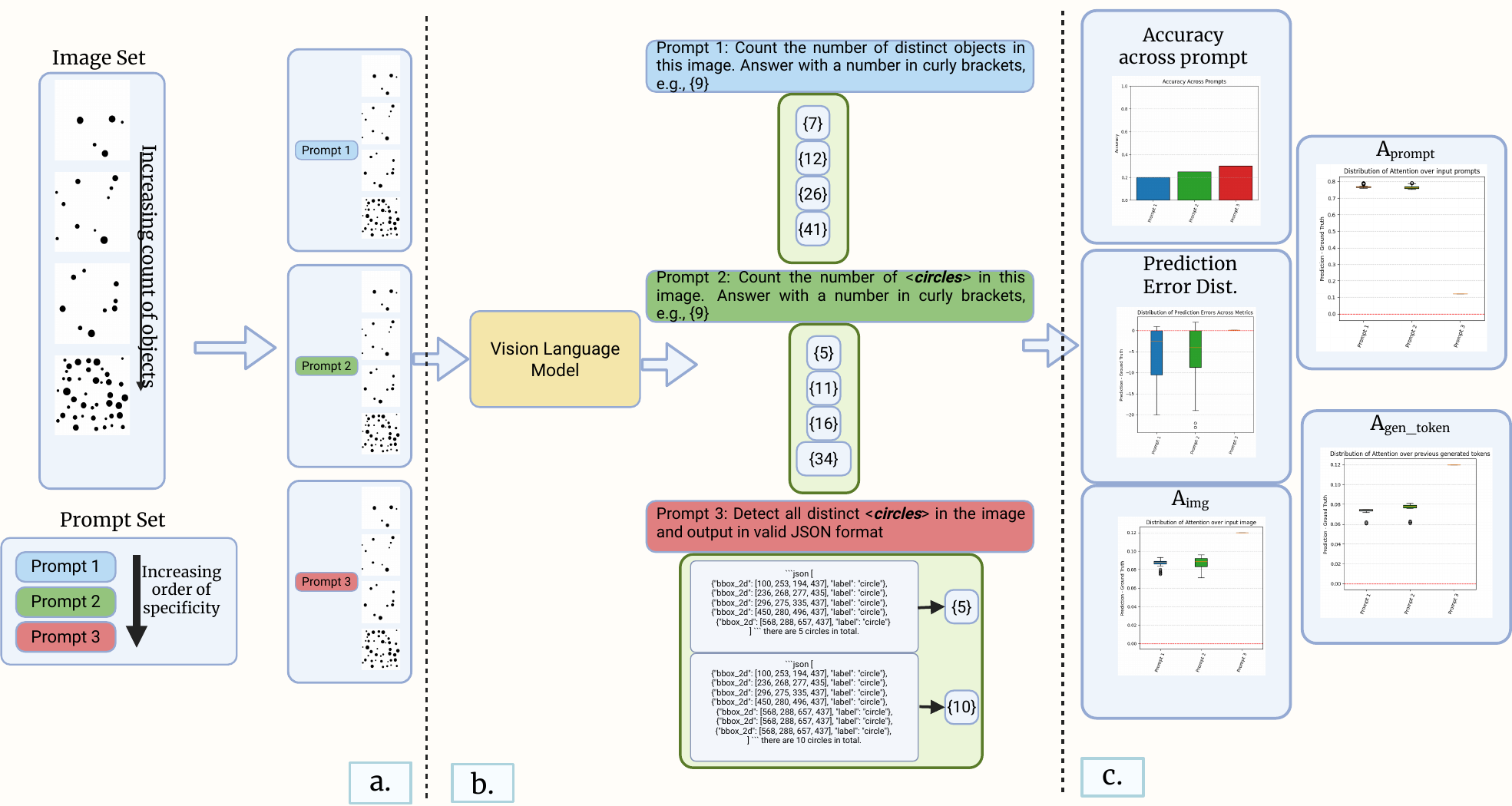}
\caption{a. We create a set of images and prompts varying features over the vision and text dimensions, in this case object shape and prompt specificity, while keeping other variables the same, in this case the object color (black), background color (white) and so on. b. We query the VLM and generate outputs and record attention values. c. We plot the error distribution, $A_{img}$, $A_{prompt}$ and $A_{gen\_token}$, which are the proportion of total attention over the image, prompt and generated tokens respectively.}
\label{method}
\end{figure}
\vspace{-0.2cm}

We create a set of synthetic datasets consisting of images and prompts to help with our examination. Each dataset within the set evaluates different aspects of the input data (both vision and text) on the VLM. Presently, this work focuses on object shape and associated prompts in increasing order of specificity to evaluate how a change in each feature affects model performance. To round out our analysis, we also utilize the altered real world data from Vo et al. \cite{vlmsarebiased} to evaluate model counting performance when using prompts in increasing order of specificity. This serves as an extension to their work, where we evaluate the effect of prompt guidance on VLM performance and its internal state. 

The goal is to measure model performance and variations in proportion of attention being ascribed to the input text tokens, vision tokens and the token being generated as output when only one aspect of the image (in this case object shape) is being varied under different prompts. Potential extensions to this problem will include more changes in image features such as object color, background color, background texture and more different versions of the prompt. This analysis focuses on handcrafted prompts.
For this analysis we focus on 2 open source VLMs, Qwen2.5 VL(both 7B and 32B Instruct variants) and \cite{qwen2.5-VL} and Kimi-VL-A3B-Instruct \cite{kimiteam2025kimivltechnicalreport} from Hugging Face \footnote{\url{https://huggingface.co/Qwen/Qwen2.5-VL-7B-Instruct}}\footnote{\url{https://huggingface.co/moonshotai/Kimi-VL-A3B-Instruct}}.


\bibliographystyle{unsrtnat} 
\bibliography{references}    

\newpage
\appendix

\section{Supplementary Material}

\subsection{Examples of prompt specificity changing model results. }
\label{sec:a1}

\begin{figure}[h!]
\centering
\includegraphics[width=0.9\linewidth]{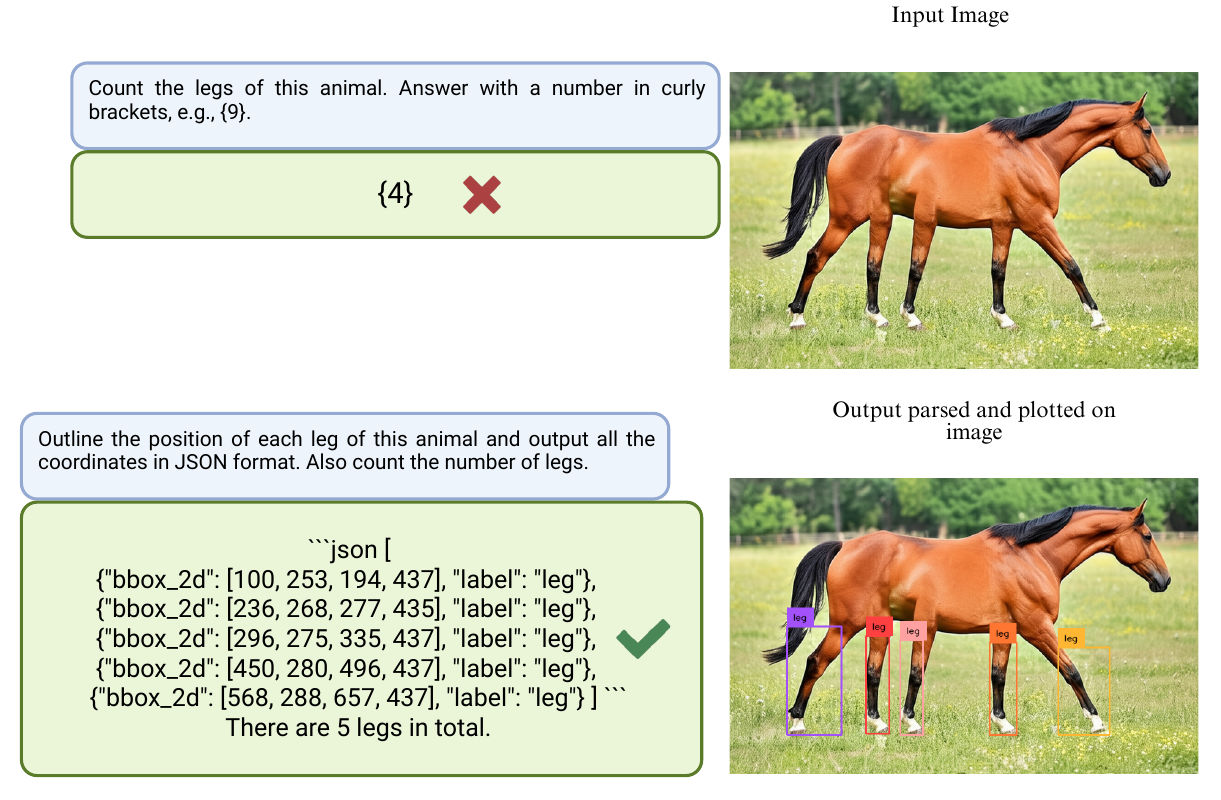}
\caption{Example of prompt specificity changing model results. Guiding the model to focus on specific areas of the image can lead to more accurate counting results. Result on querying Qwen2.5-VL-7B. Image with bounding boxes plotted using Roboflow Supervision \url{https://github.com/roboflow/supervision}.}
\label{fig:prompt_guide2}
\end{figure}

\begin{figure}[h!]
\centering
\includegraphics[width=0.9\linewidth]{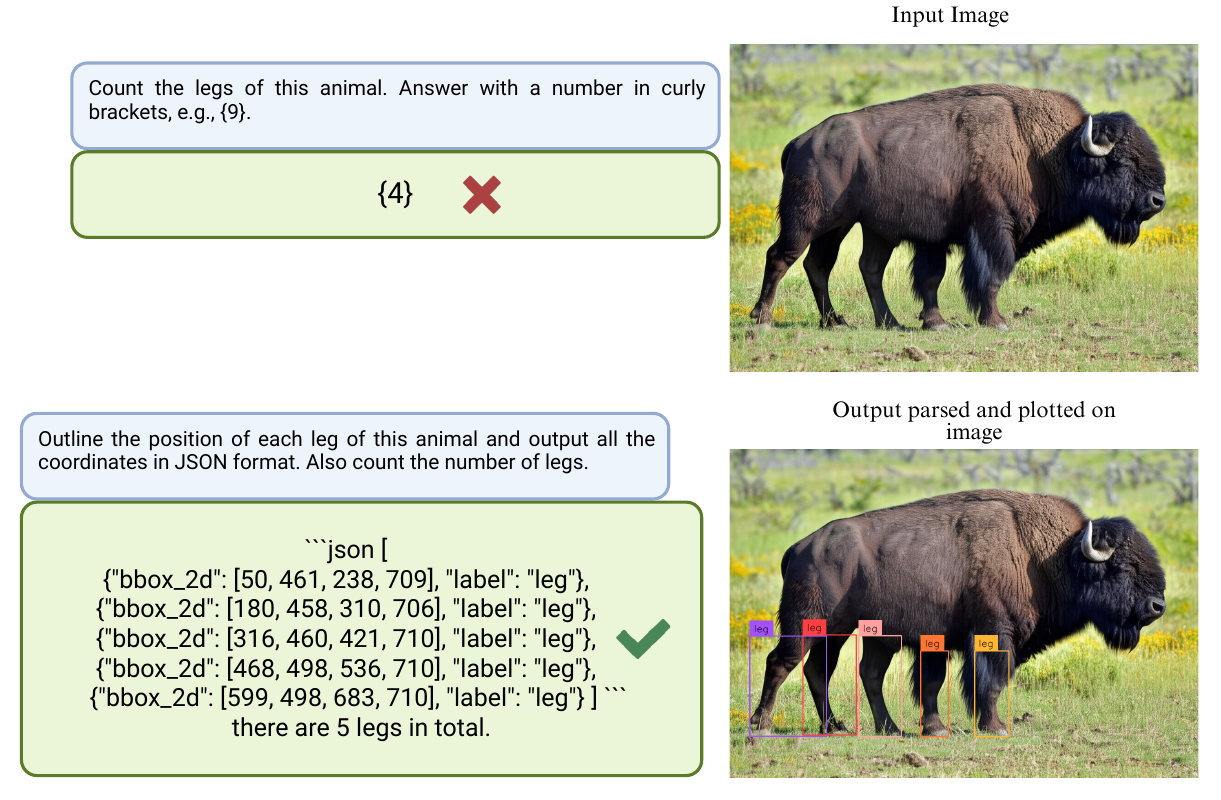}
\caption{Example of prompt specificity changing model results. Guiding the model to focus on specific areas of the image can lead to more accurate counting results. Result on querying Qwen2.5-VL-7B. Image with bounding boxes plotted using Roboflow Supervision \url{https://github.com/roboflow/supervision}.}
\label{fig:prompt_guide}
\end{figure}


\subsection{Models Used}

\paragraph{Qwen2.5-VL}

Multimodal large language model series developed by Qwen team, Alibaba Cloud, released on January 2025. The model converts images directly into tokens that mix with text tokens in one unified sequence and uses standard self-attention across the mixed sequence. Can handle both still images and video and shows good performance on tasks like document parsing and object grounding \cite{qwen2.5-VL}.

\paragraph{Kimi-VL-A3B-Instruct}

16 Billion parameter multimodal Mixture-of-Experts (MoE) model with reasoning capabilities. Shows strong performance in college-level image and video comprehension, optical character recognition (OCR), mathematical reasoning, multi-image understanding\cite{kimiteam2025kimivltechnicalreport}.

We choose these models due to their strong benchmark performance on different visual tasks and because they are open source, giving us a way to develop methods to investigate model internals \cite{qwen2.5-VL, kimiteam2025kimivltechnicalreport}. Also we choose the instruction tuned version of both these models downloaded from Hugging Face.

\subsection{Dataset Description}

Our goal with this examination framework is to measure and visualize changes in model performance and how it allocates attention to different areas of the input vision and text tokens. To do this we create synthetic images that vary over a single dimension like object shape, and do the same over the text dimension. We keep the other dimensions like object color, background color etc. the same. As an example in Table \ref{tab:dimensions} we show the values across each dimension. We also show which values we chose to vary for this analysis. 



\begin{table}[h]
\centering
\caption{For this preliminary analysis we only vary single dimension each on the image and prompt axes. \colorbox[HTML]{FE996B}{Red} signifies changed values, while \colorbox[HTML]{DAE8FC}{blue} signifies values kept constant. Refer to Figure \ref{fig:result_qwen7b} for an example.}
\label{tab:dimensions}
\begin{tabular}{ll}
\toprule
\textbf{Image Axis}                  & \textbf{Prompt Axis} \\ \midrule
\rowcolor[HTML]{FE996B} 
Object Shape                                & Prompt Guidance         \\
\cellcolor[HTML]{DAE8FC}{Object Color}        &     \cellcolor[HTML]{DAE8FC}{Prompt Semantics}                       \\
\cellcolor[HTML]{DAE8FC}Object Texture      &                            \\ 
\cellcolor[HTML]{DAE8FC}Background Color   &                            \\
\cellcolor[HTML]{DAE8FC}Background Texture &                            \\
\cellcolor[HTML]{DAE8FC}Image Dimensions &                            \\
\cellcolor[HTML]{DAE8FC}Image File Type &                            \\
\bottomrule
\end{tabular}
\end{table}

\subsubsection{Synthetic Images}

\begin{figure}[h]
\centering
\includegraphics[width=\linewidth]{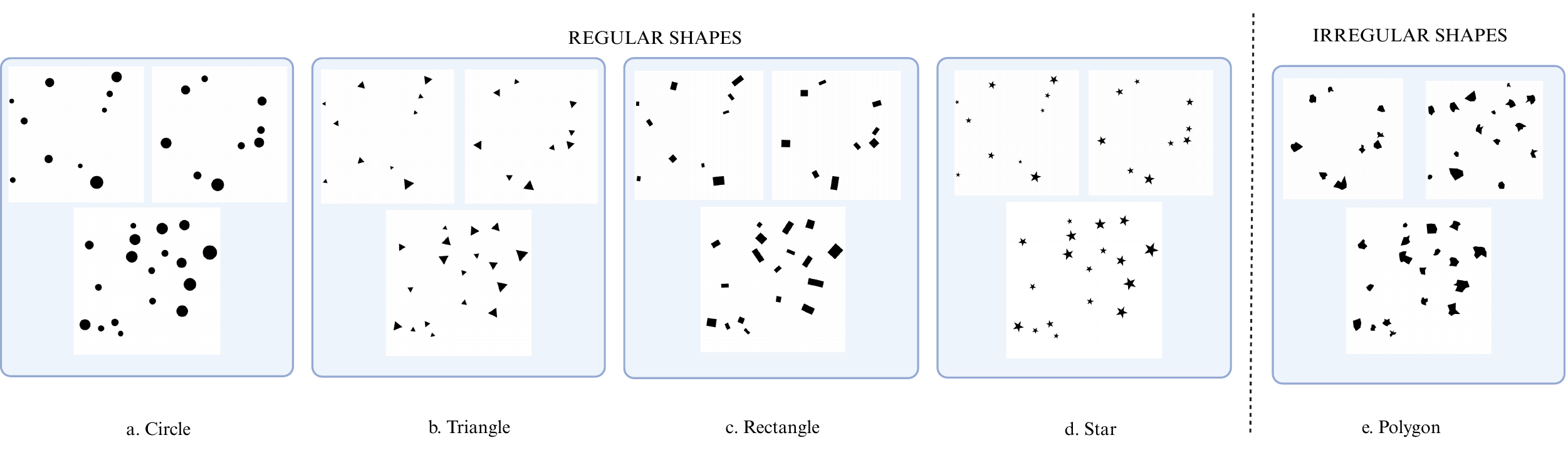}
\caption{Example Images from our synthetic dataset}
\label{fig:shape_eg}
\end{figure}

For this analysis we vary the shape of the object keeping the other variables the same. We choose the following object shapes for this analysis: circle, triangle, rectangle, star and polygons. The number of objects created are in buckets of 1-10, 10-20, 20-30,30-40,40-50 with each bucket represented equally. We create 50 images each for all object shapes. All objects are in the same location with just the object shape and count changing. We also take care keep the image dimensions and image file type constant. Refer to Figure \ref{fig:shape_eg} for an illustration.

\subsubsection{Altered Real World Images}

We utilize images from Vo et al. \cite{vlmsarebiased} as an example of real world data for our evaluation. We use the Animals and Flag Stars datasets. The `Animals' dataset contains altered images of various animals at different image sizes with each animal sporting a natural looking extra leg, designed to fool current VLMs. The `Flags' dataset contains flags from countries that have stars in their flags and place one extra star within the image.

\subsubsection{Prompts}

For this analysis, we curate handcrafted prompts. Due to constraints on time and computational resources, we select three prompts with progressively increasing specificity for each image class. In other words, each subsequent prompt is more precise in its formulation than the preceding one. Future work will automating prompt design with the help of Large Language Models (LLMs). Refer to Table \ref{tab:prompts} for details.

\begin{table}[h]
\centering
\caption{Prompts used for each task.}
\label{tab:prompts}
\resizebox{\columnwidth}{!}{%
\begin{tabular}{llp{8cm}p{8cm}}
\hline
\textbf{Object Class}                         & \textbf{\#} & \textbf{Prompt}                                                                                                                & \textbf{Comments}                                                                                  \\ \hline
\multirow{3}{*}{Synthetic Images}             & Prompt 1    & Count the number of distinct objects in this image. Answer with a number in curly brackets, e.g., \{9\}                        & Does not specify what type of object                                                               \\ \cline{2-4} 
                                              & Prompt 2    & Count the number of \textless{}shape\textgreater in this image.  Answer with a number in curly brackets, e.g., \{9\}           & Replace \textless{}shape\textgreater with "circles", "triangles, "rectangles","stars", "polygons". Explicit focus on object\\ \cline{2-4} 
                                              & Prompt 3    & Detect all distinct\textless{}shape\textgreater in the image and output valid JSON format                                      & More precise and explicit focus on object.                                                     \\ \hline
\multirow{3}{*}{Real World Data (Animal) from \cite{vlmsarebiased}}     & Prompt 1    & Count the legs of this animal. Answer with a number in curly brackets, e.g., \{9\}.                                            & Does not specify what type of animal. Same as Vo et al.\cite{vlmsarebiased}.                                                       \\ \cline{2-4} 
                                              & Prompt 2    & Outline the position of each leg of this animal and output all the coordinates in JSON format. Also count the number of legs.  & Explicitly asks to focus on Legs.                                                                  \\ \cline{2-4} 
                                              & Prompt 3    & Outline the position of each feet of this animal and output all the coordinates in JSON format. Also count the number of legs. & Explicitly asks to focus on Feet.                                                                  \\ \hline
\multirow{3}{*}{Real World Data (Flag Stars) from \cite{vlmsarebiased}} & Prompt 1    & Count the number of objects in this image.  Answer with a number in curly brackets, e.g., \{9\}                                & Does not specify what object                                                                       \\ \cline{2-4} 
                                              & Prompt 2    & How many stars are there in this flag? Answer with a number in curly brackets, e.g., \{9\}.                                    & Specifies the type of object to count. Same as Vo et al.\cite{vlmsarebiased}.                                                        \\ \cline{2-4} 
                                              & Prompt 3    & Outline the position of each star in this image and output all the coordinates in JSON format. Also count the number of stars. & Explicit focus on object and activates object detection.                                           \\ \hline
\end{tabular}%
}
\end{table}

\subsection{Calculation of attention proportion}

To measure how the model distributes attention across different parts of the input (i.e. visual tokens, input prompt tokens, and previously generated tokens), we compute an attention proportion for each generated token after aggregating across all layers and heads. For a generated token $g$, every layer $l$ and head $h$ produces an attention vector over the context tokens (visual and input prompt tokens and all previously generated tokens), which are first averaged across heads within each layer and then across all layers to yield a single vector $A_{g} \in \mathbb{R}^{S+g-1} $, where $S$ is the input prompt size (i.e. image and input text). This vector is then partitioned into three regions, including "vision tokens", "input prompt text tokens", and "previously generated tokens". We finally report the mean proportion across all generated tokens for each region, denoted by $A_{img}$, $A_{prompt}$, $A_{gen\_token}$ respectively, which together sum to one.

\subsection{Results}

We report and visualize the following metrics:

\begin{enumerate}
    \item Object counting accuracy over different images grouped by prompt.
    \item Prediction Error, defined by taking the difference between the ground truth object count and the predicted count over all shapes/data grouped by prompt
    \item Distribution of $A_{img}$ over all shapes/data grouped by prompt.
    \item Distribution of $A_{prompt}$ over all shapes/data grouped by prompt.
    \item Distribution of $A_{gen\_token}$ over all shapes/data grouped by prompt.
\end{enumerate}

We report these metrics over Qwen2.5-VL-7B, Qwen2.5-VL-32B and Kimi-VL-A3B. You can find the results for our synthetic dataset in Figures \ref{fig:result_qwen7b},\ref{fig:result_qwen32b} and \ref{fig:res_kimi}. For our results on the data from Vo et al. \cite{vlmsarebiased} refer to Figures \ref{fig:res_rwd_qwen7b} and \ref{fig:res_rwd_qwen32b}.

We also visualize how the prediction error changes over increasing number of object counts which can be found at Figures \ref{fig:errors_qwen7b},\ref{fig:errors_qwen32b} and \ref{fig:errors_kimi}.

\subsection{Discussion}

\begin{figure}[h]
            \centering
            
                \begin{tabular}{cc}
                     \includegraphics[width=0.5\linewidth]{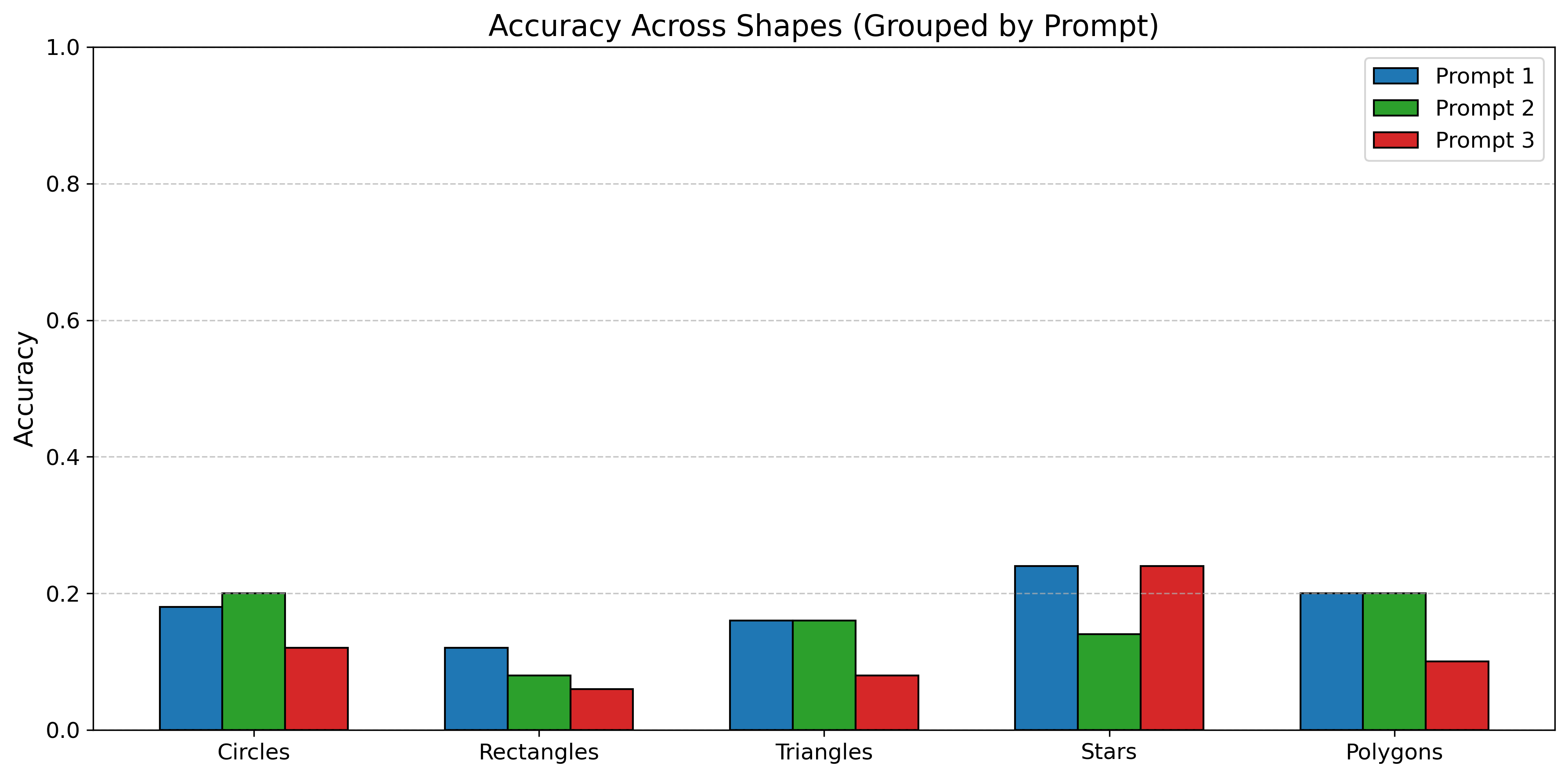}
                    & \includegraphics[width=0.5\linewidth]{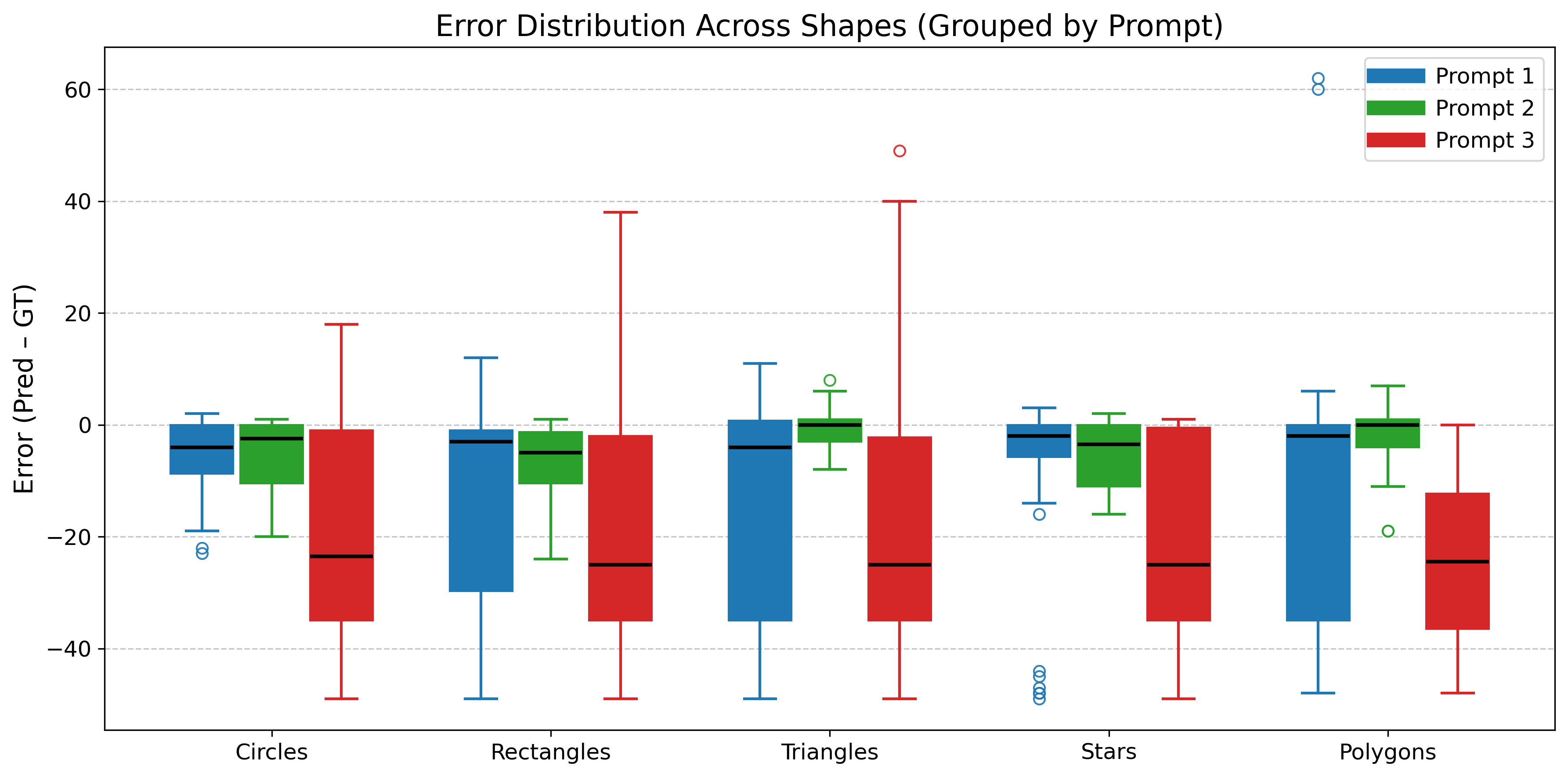}\\[-4pt]
                    \includegraphics[width=0.5\linewidth]{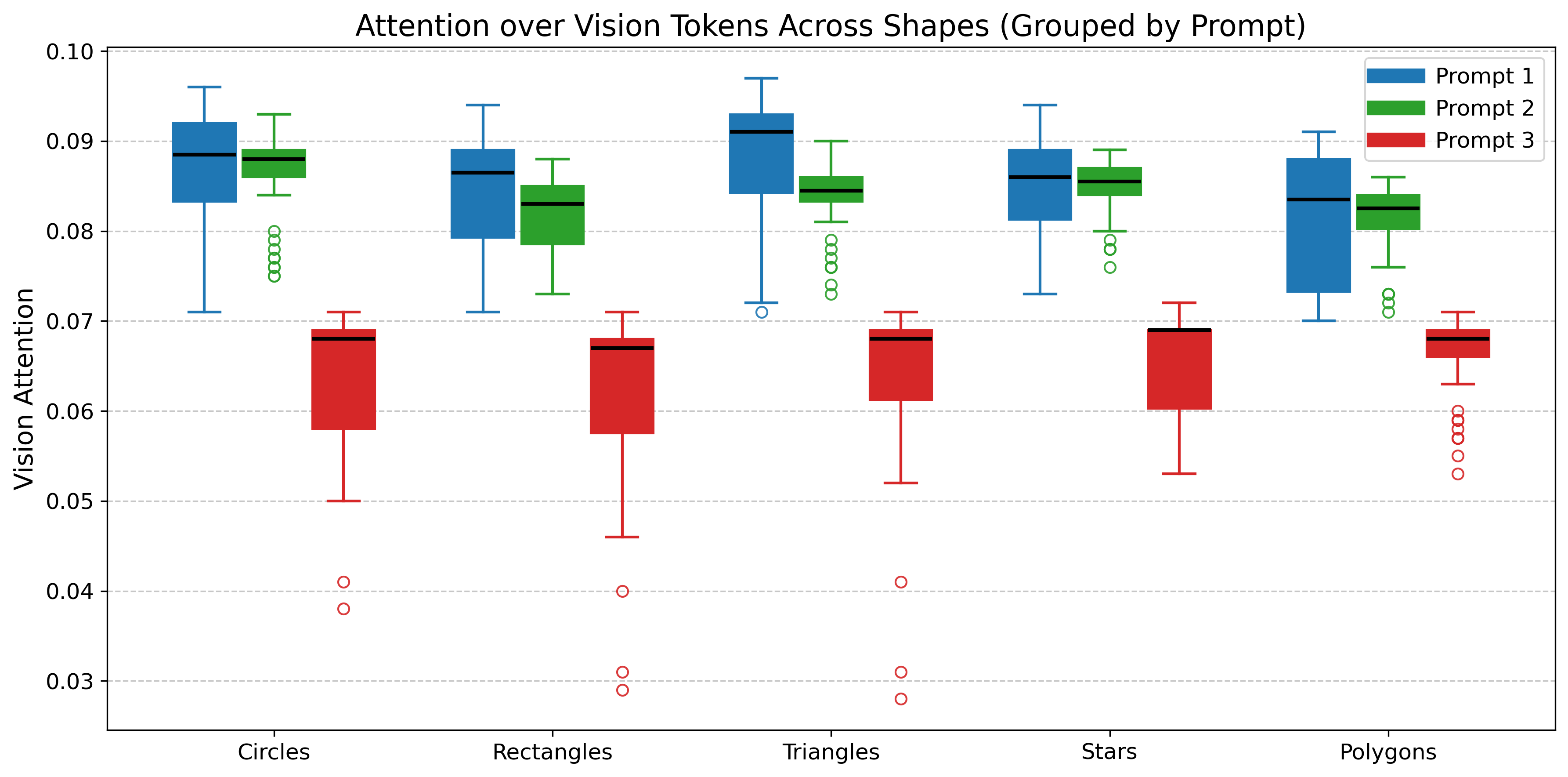}
                    & \includegraphics[width=0.5\linewidth]{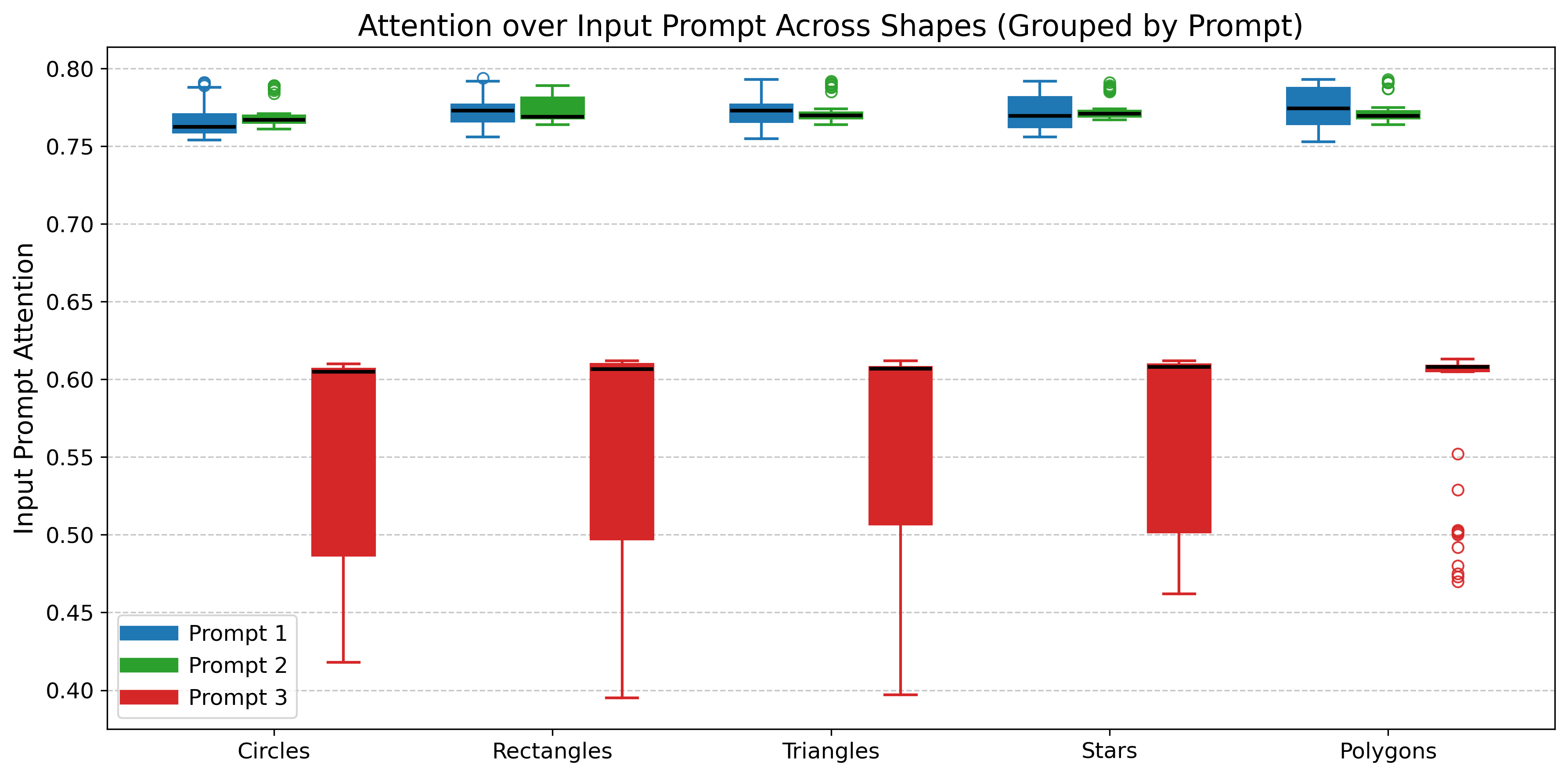}\\[-4pt]
                    \multicolumn{2}{c}{ \includegraphics[width=0.5\linewidth]{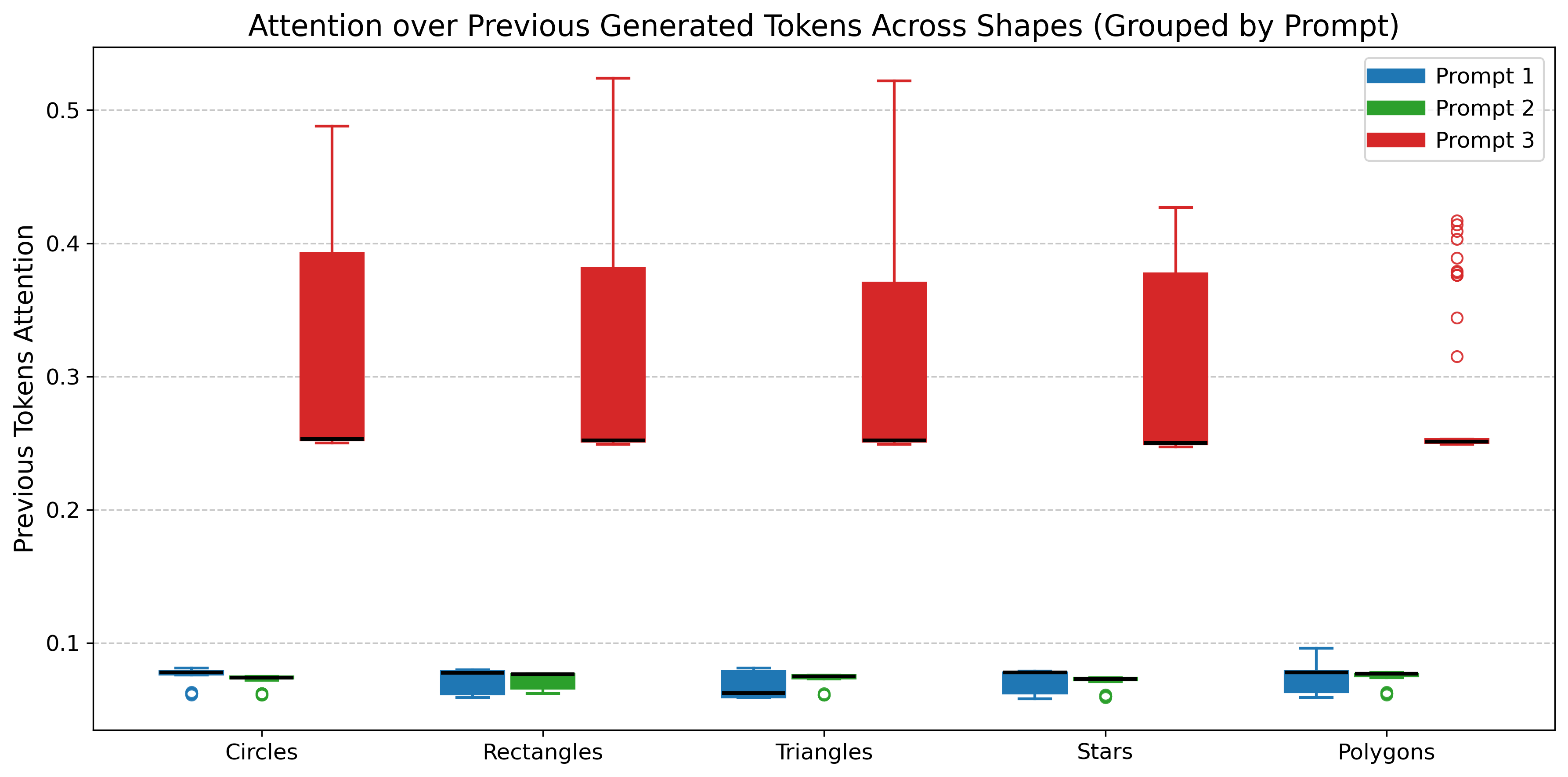}}
                    
                    \\
                \end{tabular}%
            \caption{Results on Qwen-2.5-VL-7B}
            \label{fig:result_qwen7b}
\end{figure}
\begin{figure}[h]
            \centering
            
                \begin{tabular}{cc}
                     \includegraphics[width=0.5\linewidth]{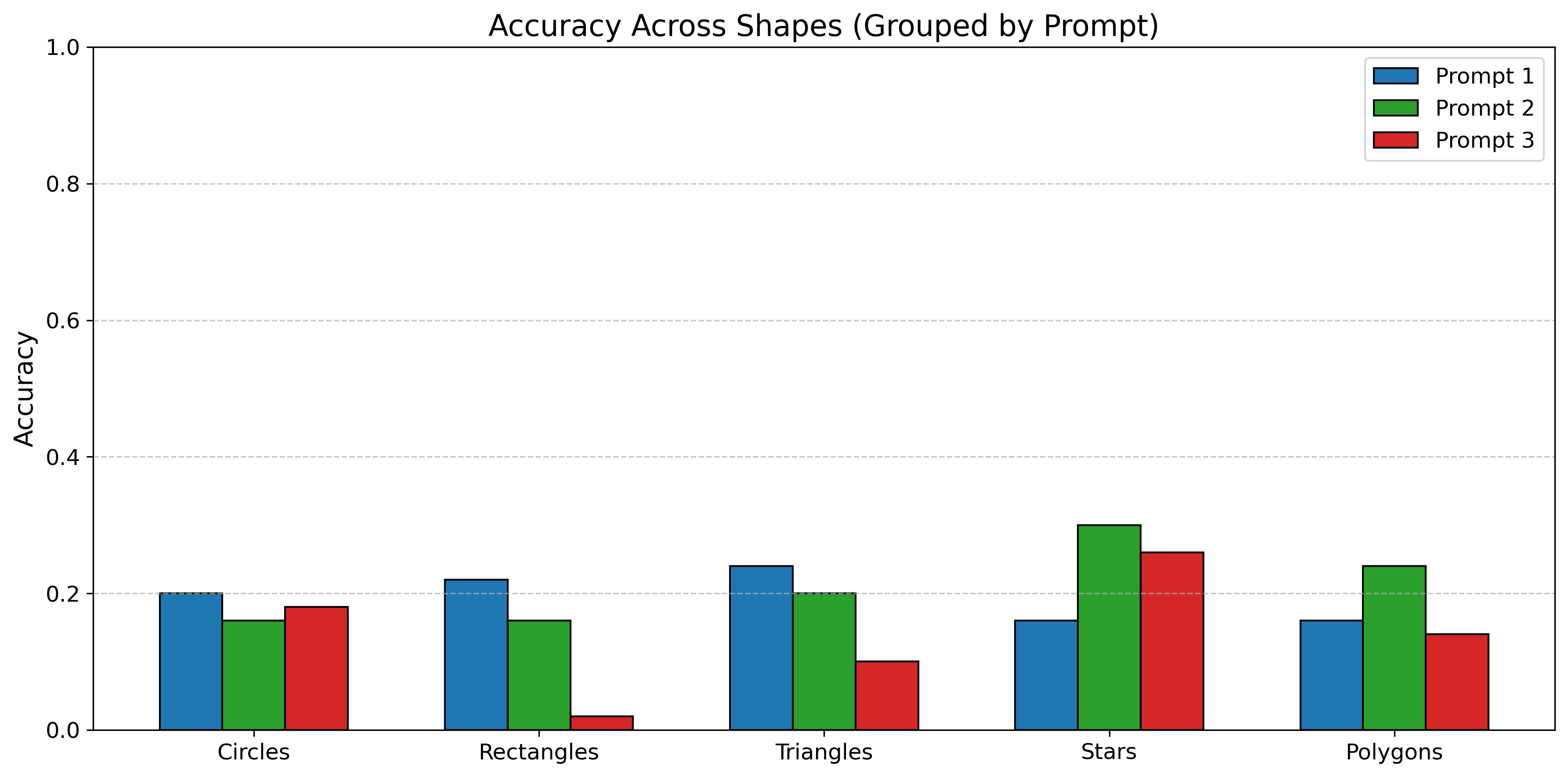}
                    & \includegraphics[width=0.5\linewidth]{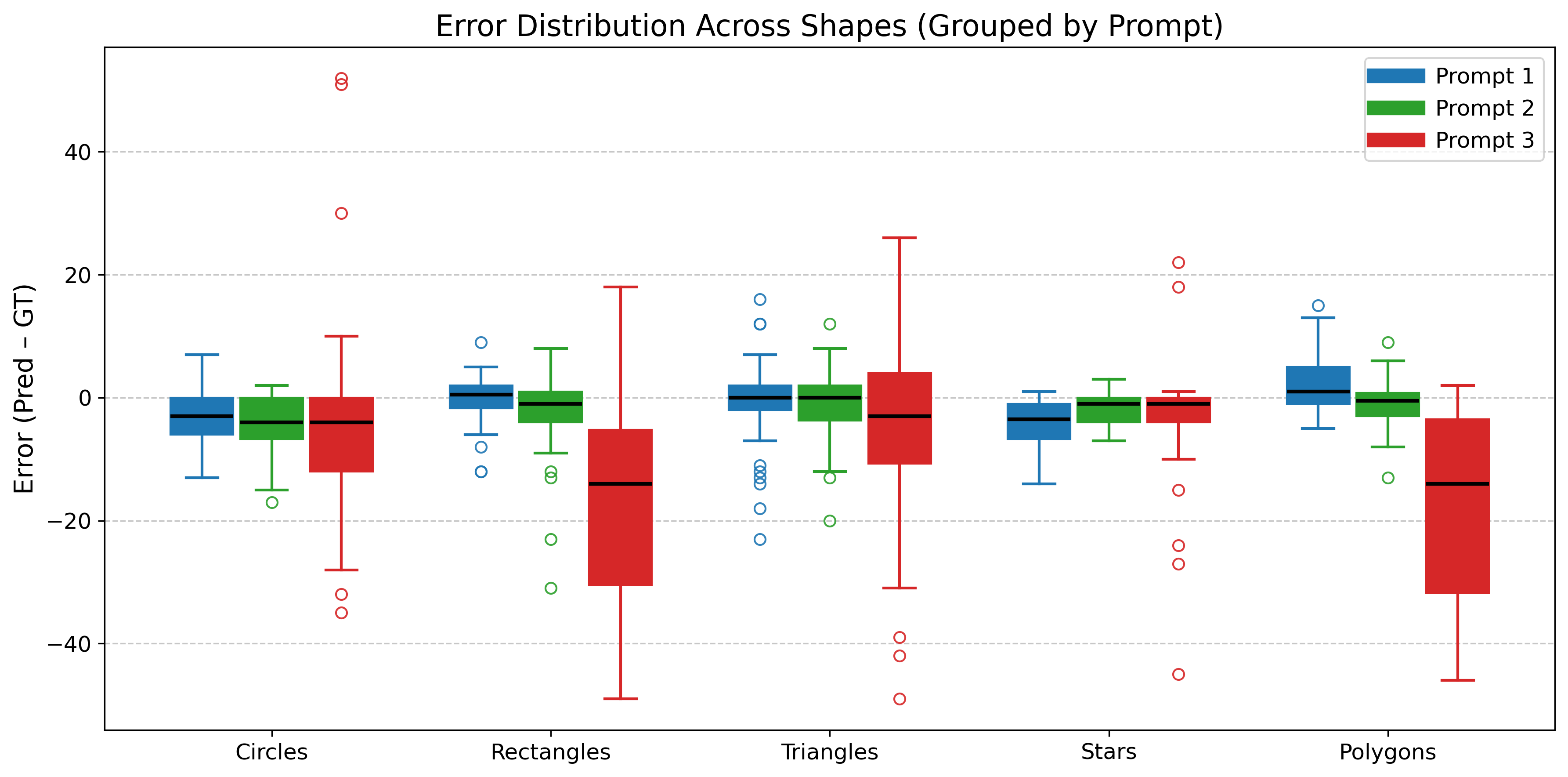}\\[-4pt]
                    \includegraphics[width=0.5\linewidth]{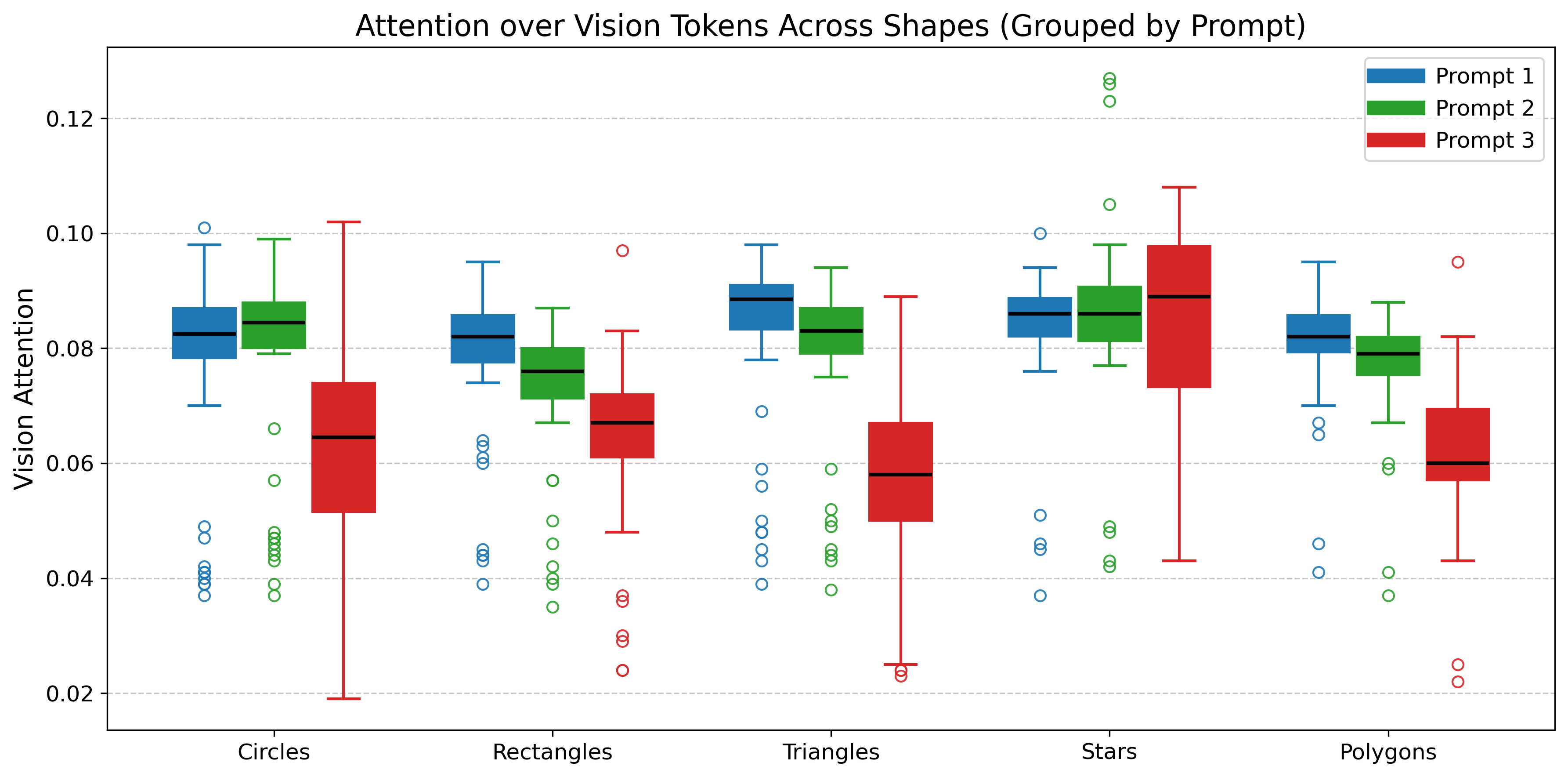}
                    & \includegraphics[width=0.5\linewidth]{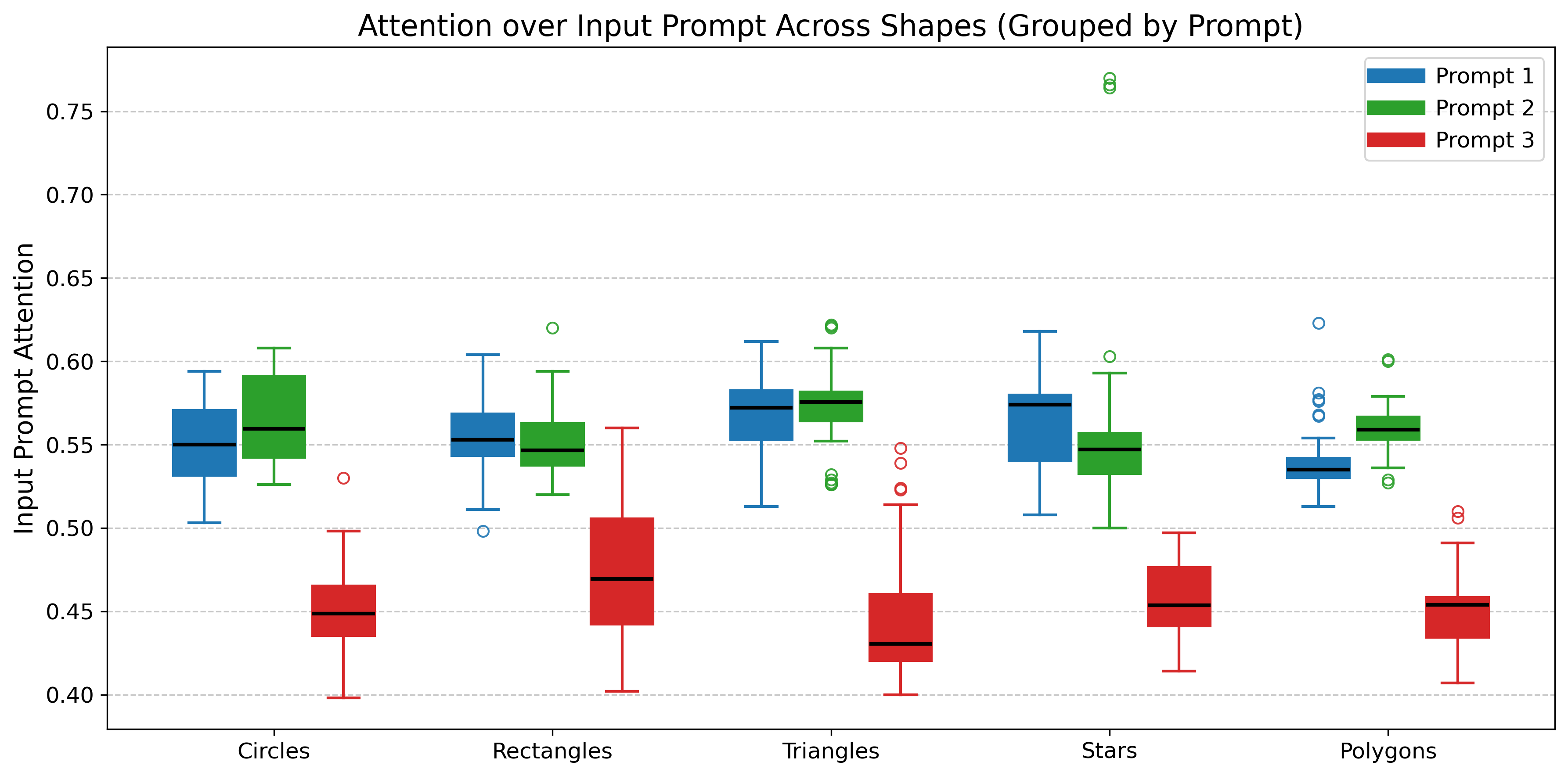}\\[-4pt]
                    \multicolumn{2}{c}{ \includegraphics[width=0.5\linewidth]{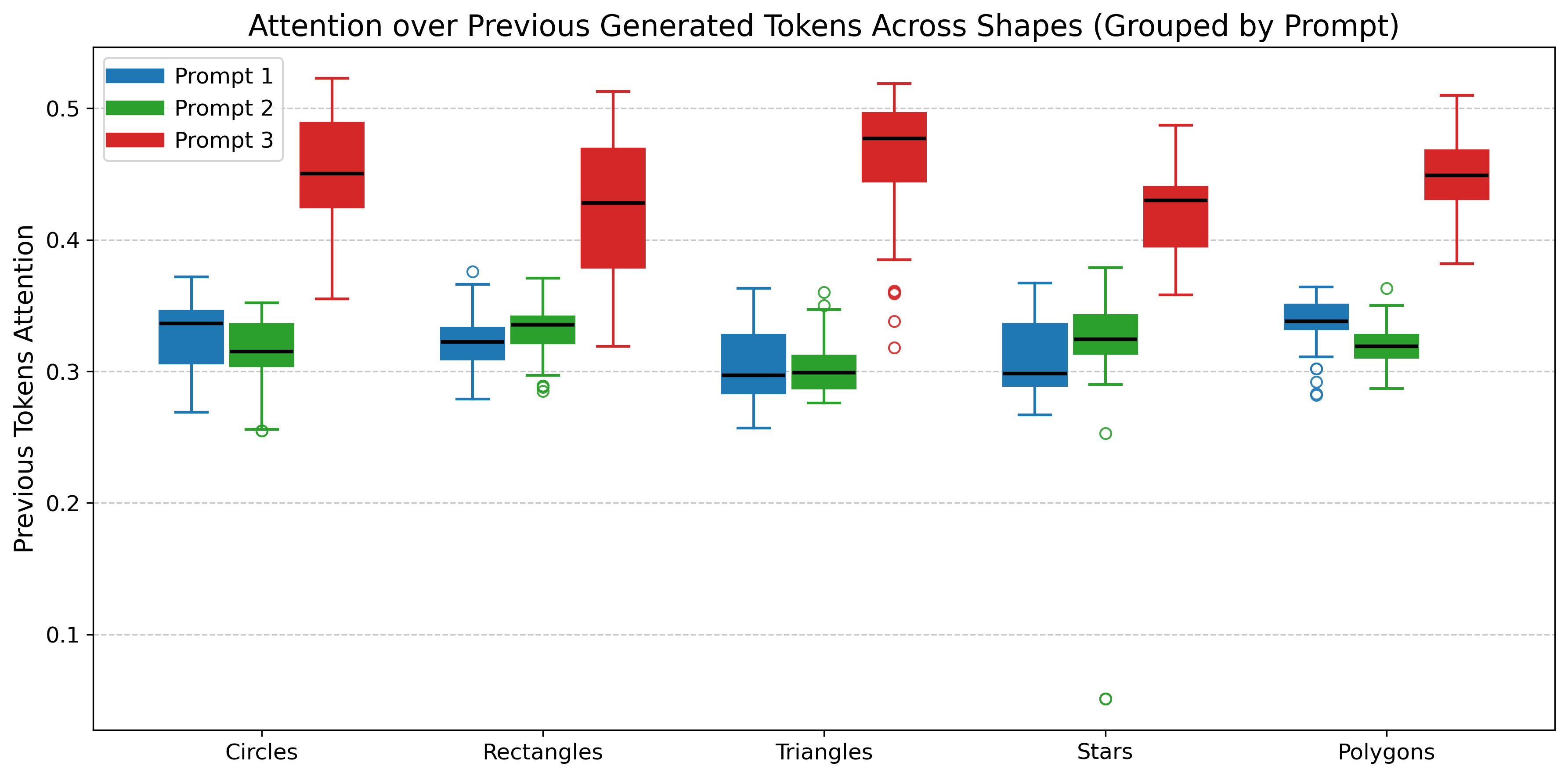}}
                    
                    \\
                \end{tabular}%
            \caption{Results on Qwen-2.5-VL-32B}
            \label{fig:result_qwen32b}
\end{figure}
\begin{figure}[h]
            \centering
            
                \begin{tabular}{cc}
                     \includegraphics[width=0.5\linewidth]{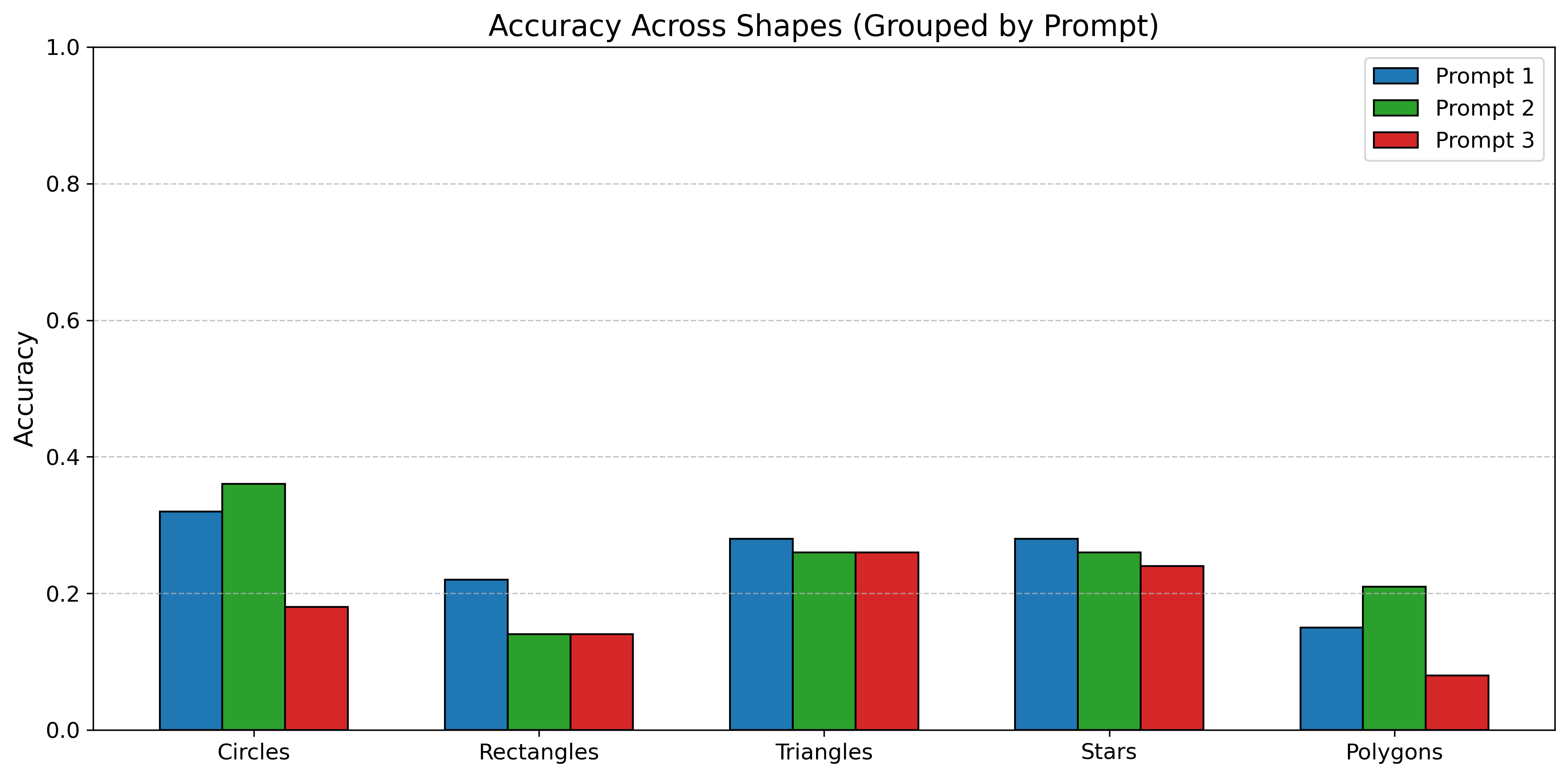}
                    & \includegraphics[width=0.5\linewidth]{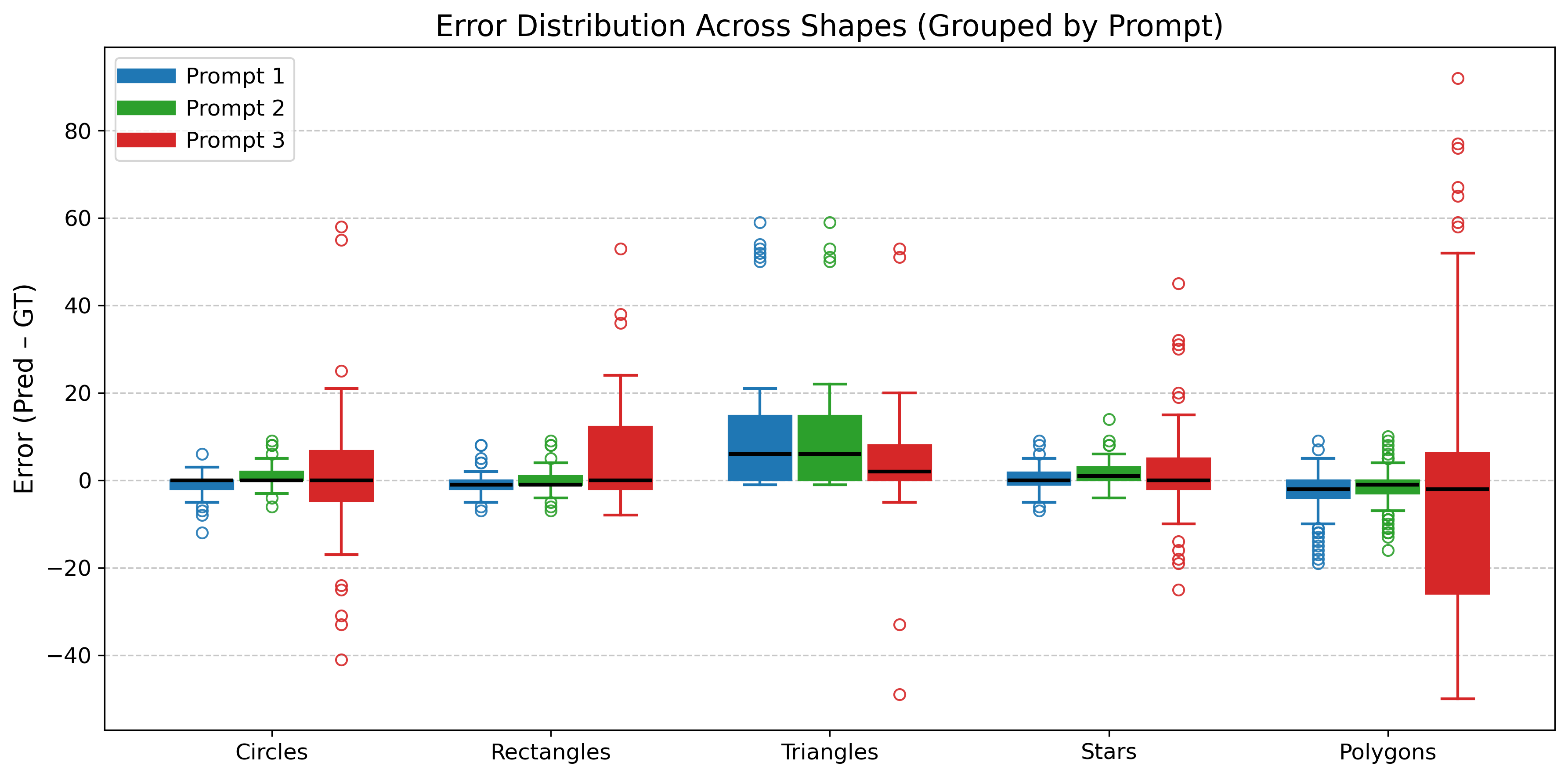}\\[-4pt]
                    \includegraphics[width=0.5\linewidth]{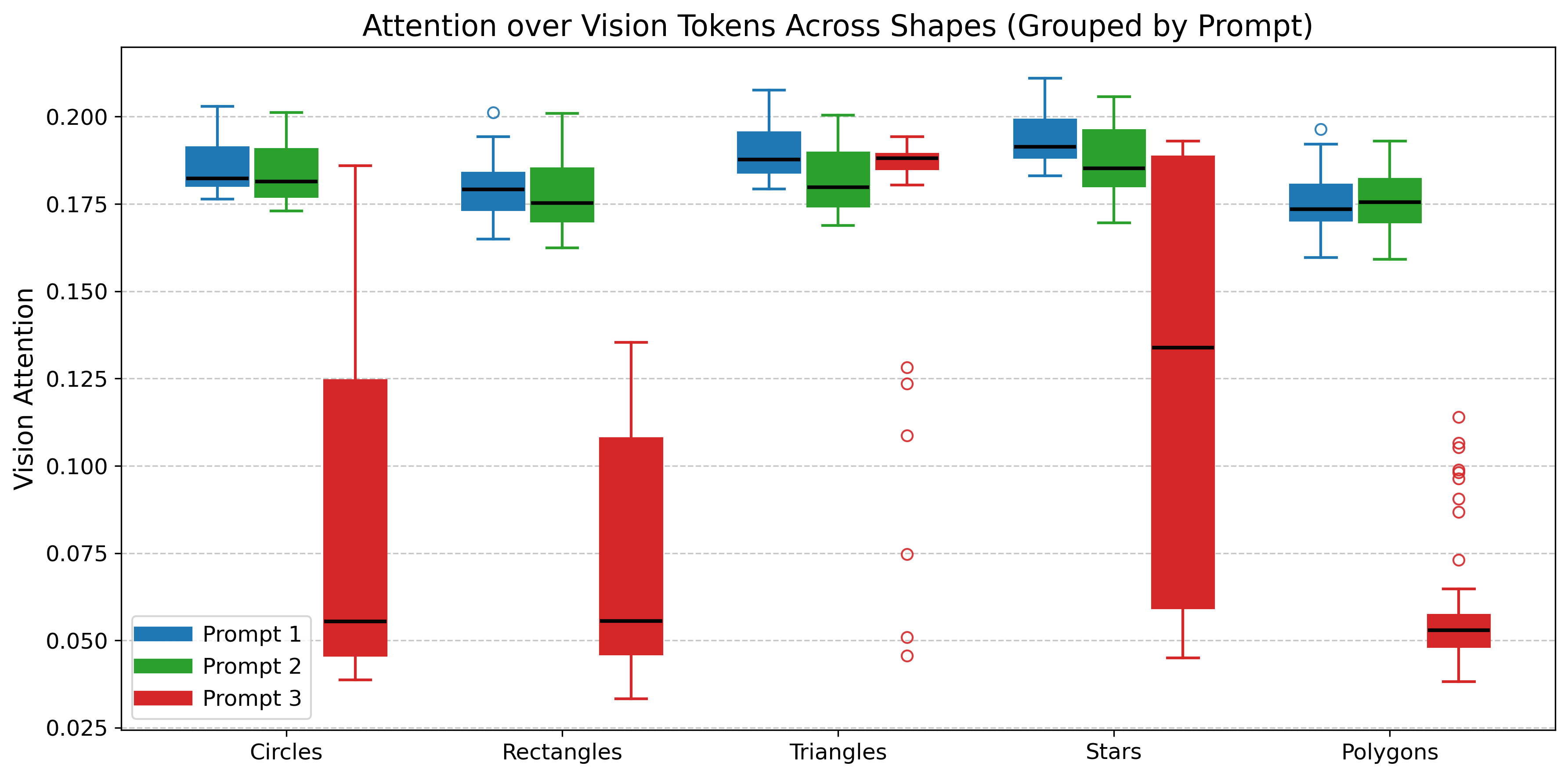}
                    & \includegraphics[width=0.5\linewidth]{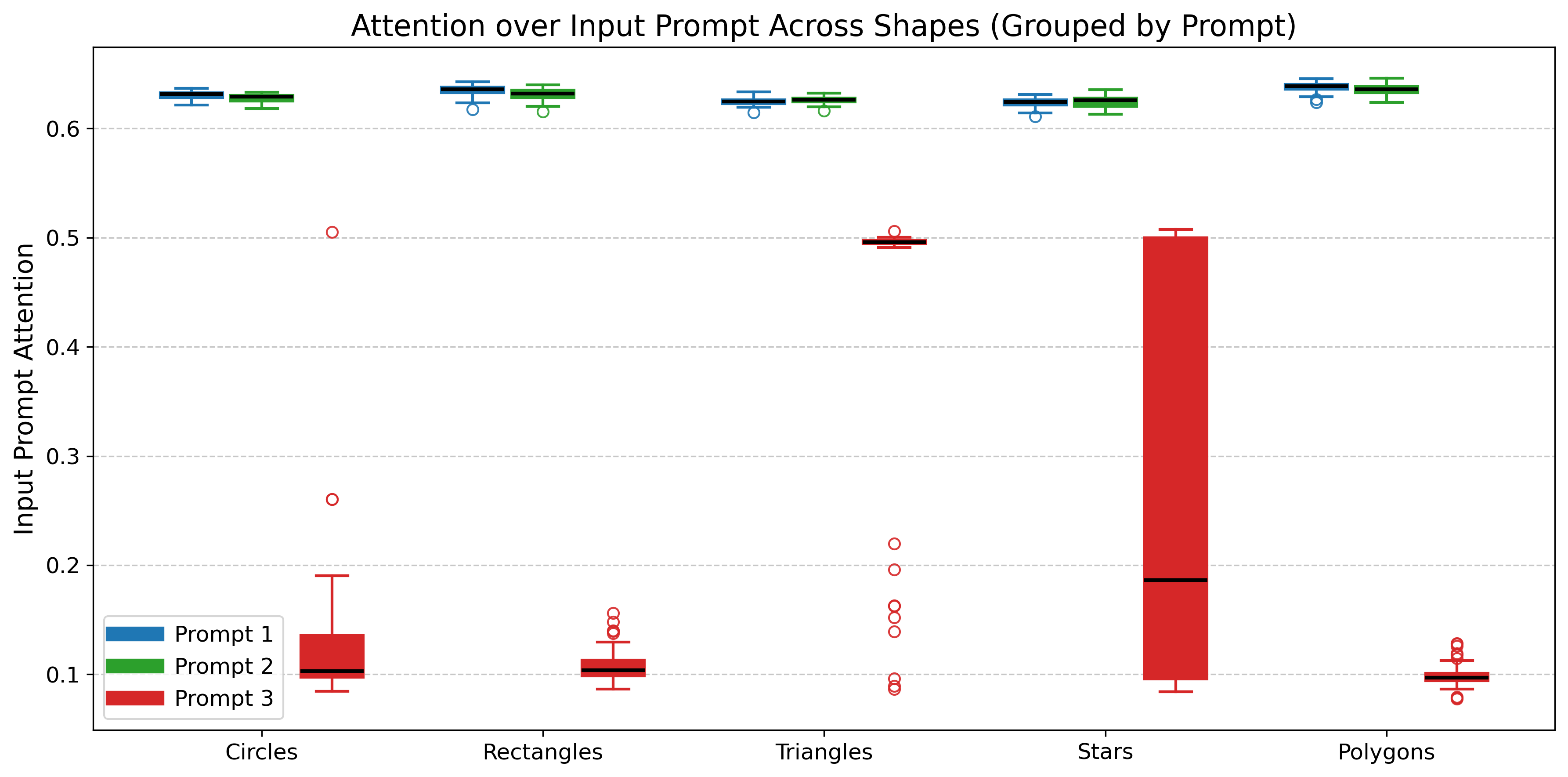}\\[-4pt]
                    \multicolumn{2}{c}{ \includegraphics[width=0.5\linewidth]{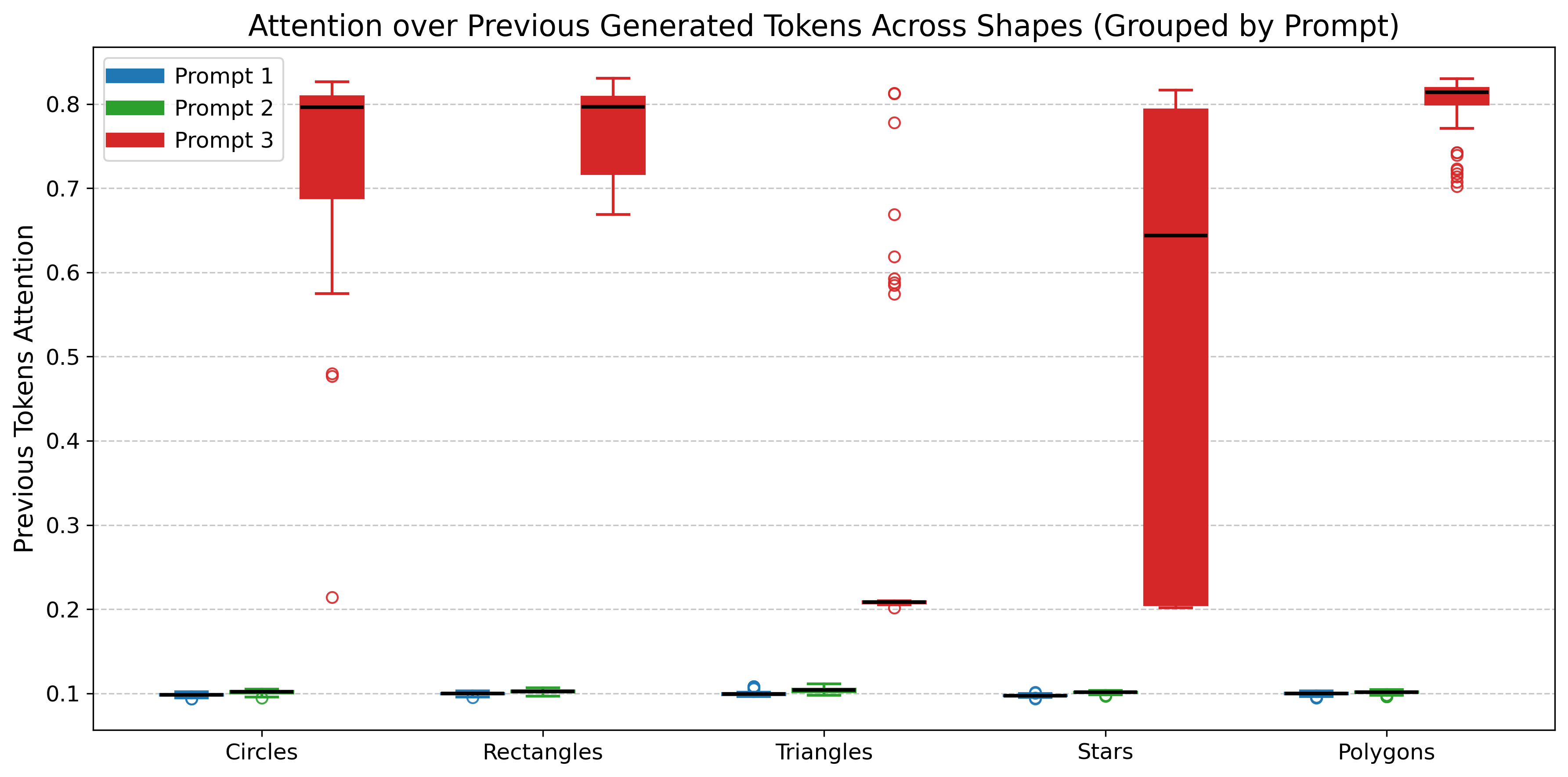}}
                    
                    \\
                \end{tabular}%
            \caption{Results on Kimi-VL-A3B}
            \label{fig:res_kimi}
\end{figure}
\begin{figure}[h]
            \centering
            
                \begin{tabular}{cc}
                     \includegraphics[width=0.5\linewidth]{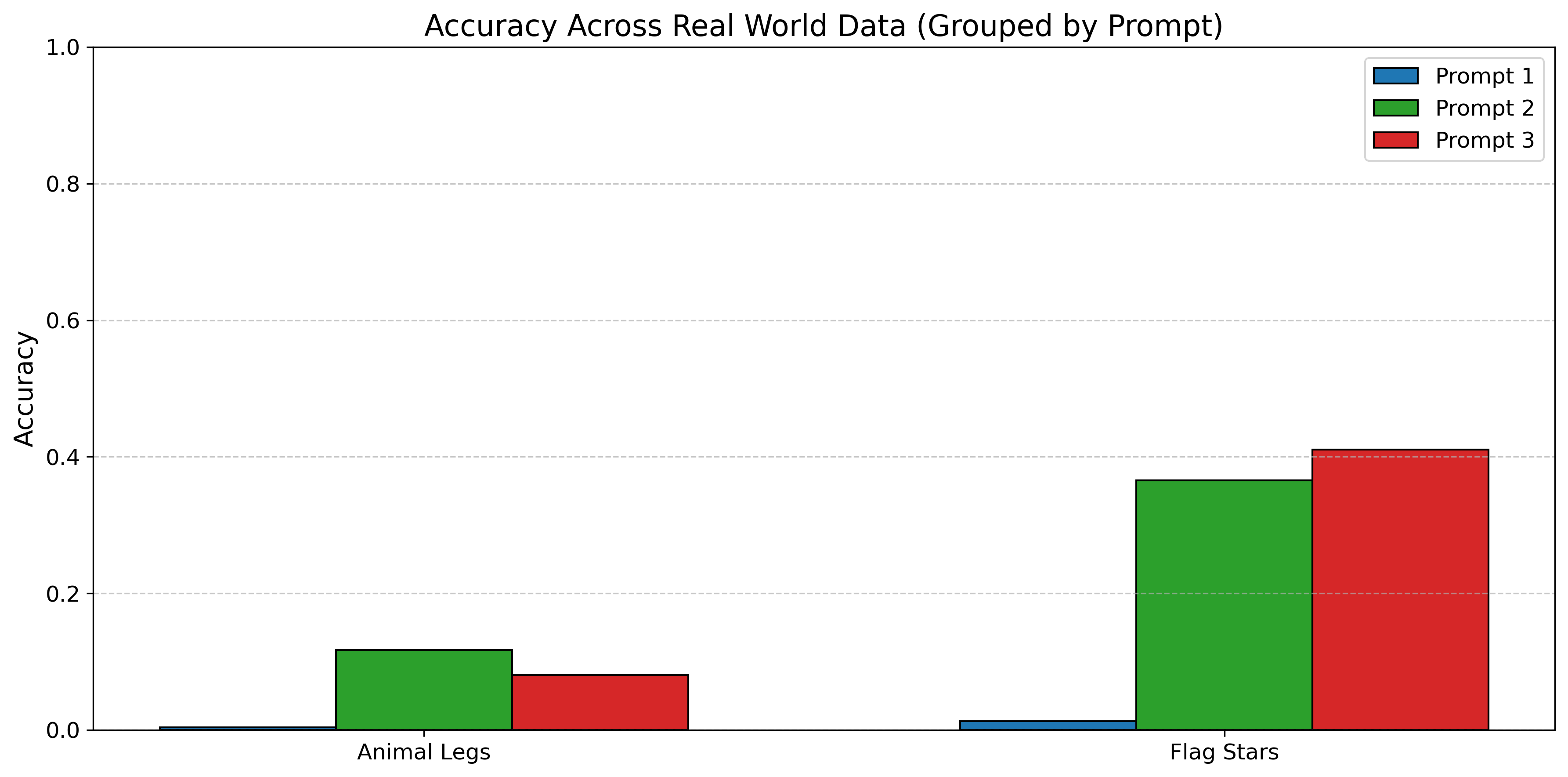}
                    & \includegraphics[width=0.5\linewidth]{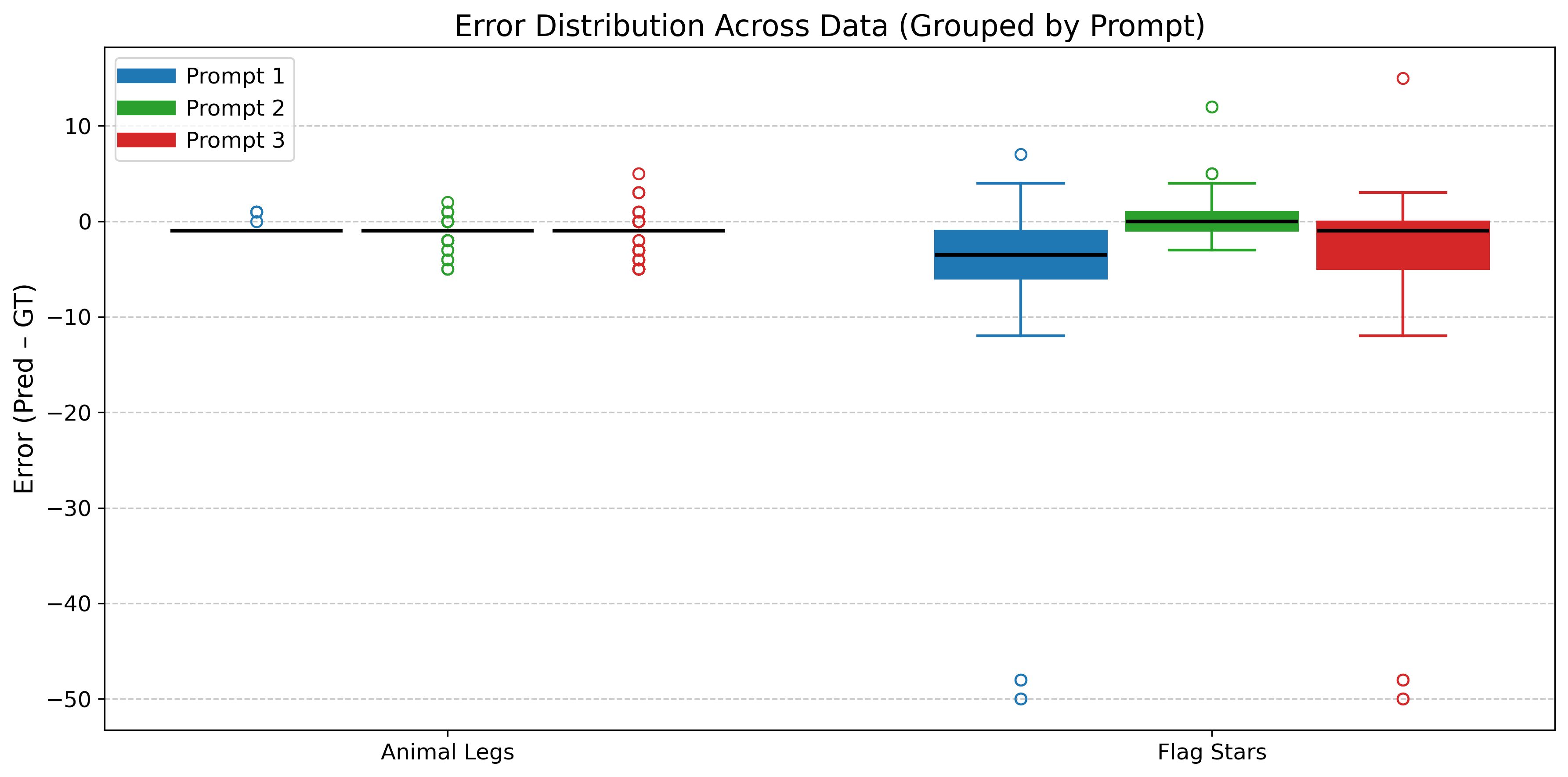}\\[-4pt]
                    \includegraphics[width=0.5\linewidth]{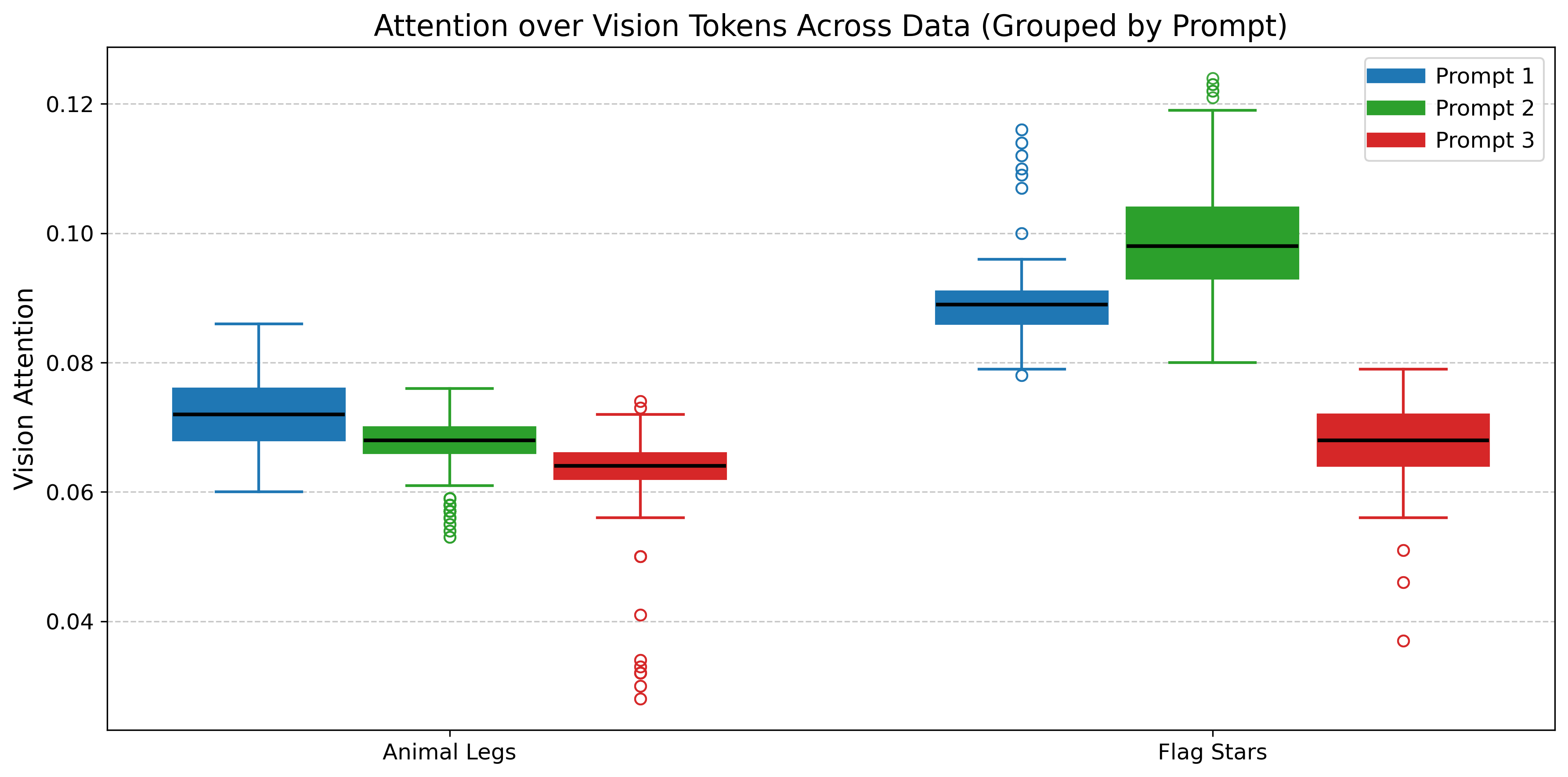}
                    & \includegraphics[width=0.5\linewidth]{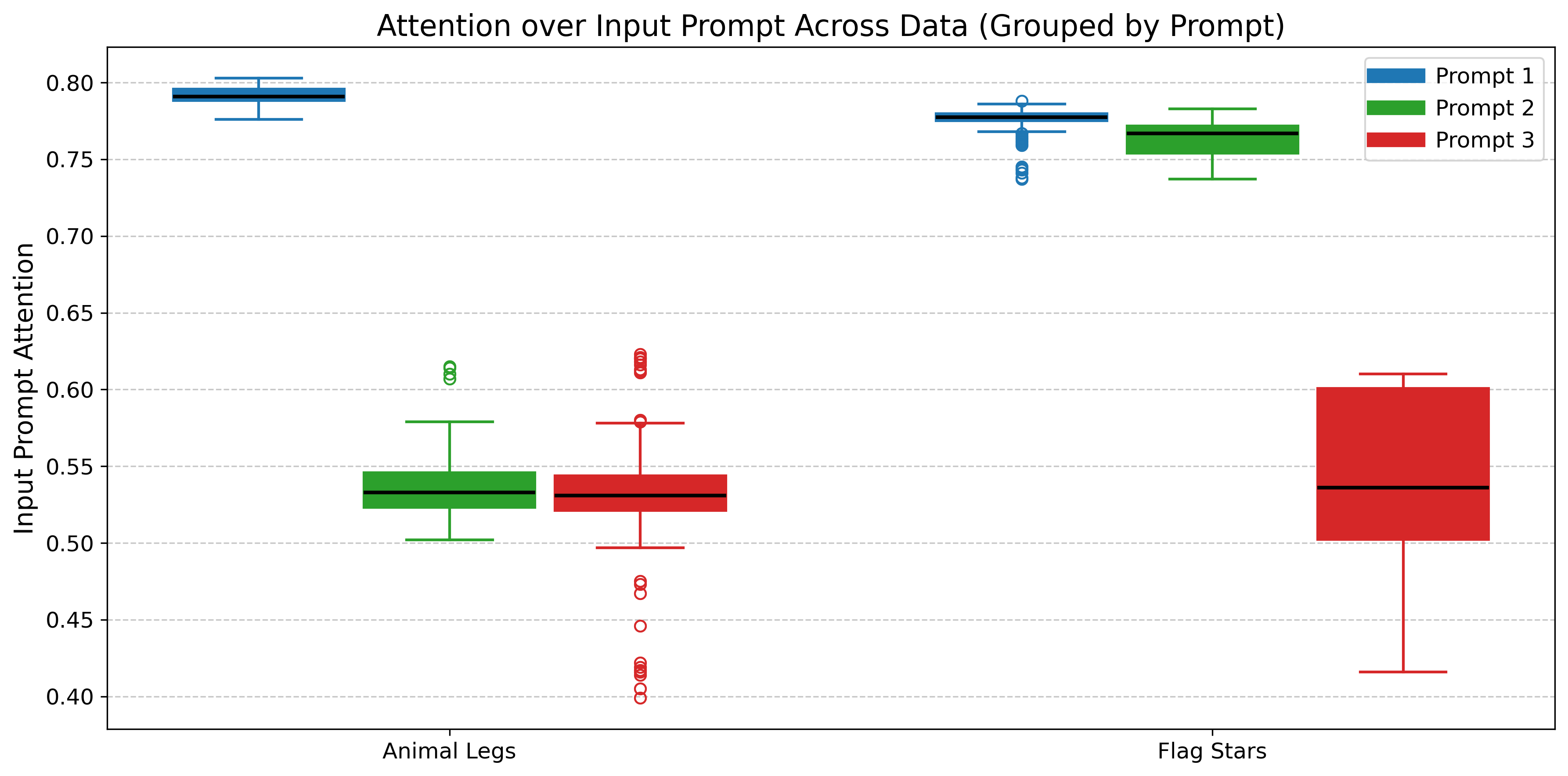}\\[-4pt]
                    \multicolumn{2}{c}{ \includegraphics[width=0.5\linewidth]{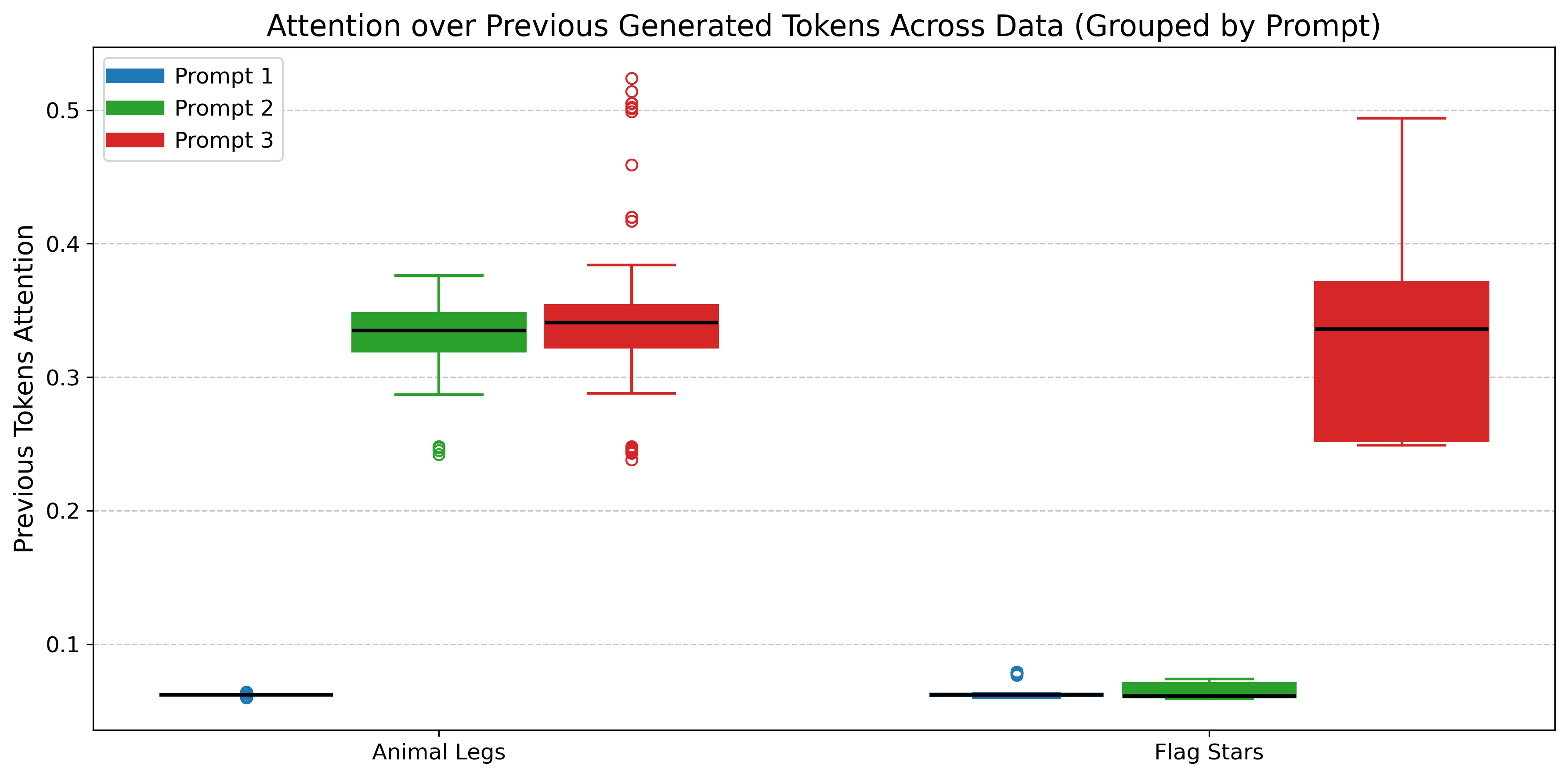}}
                    
                    \\
                \end{tabular}%
        \caption{Results on Qwen-2.5-VL-7B for data from `VLMs are Biased' paper from Vo et al. \cite{vlmsarebiased}}
        \label{fig:res_rwd_qwen7b}
\end{figure}
\begin{figure}[h]
            \centering
            
                \begin{tabular}{cc}
                     \includegraphics[width=0.5\linewidth]{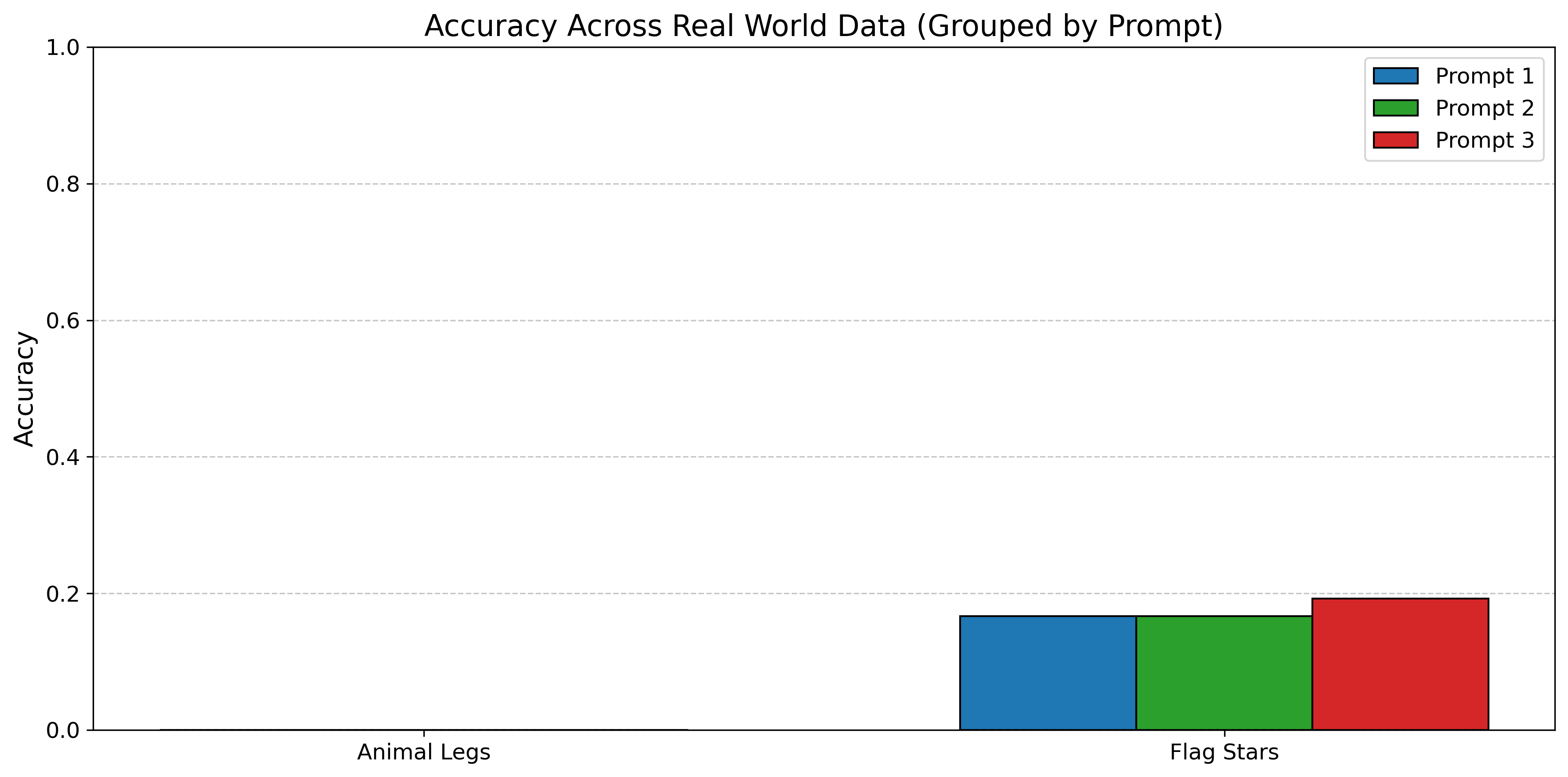}
                    & \includegraphics[width=0.5\linewidth]{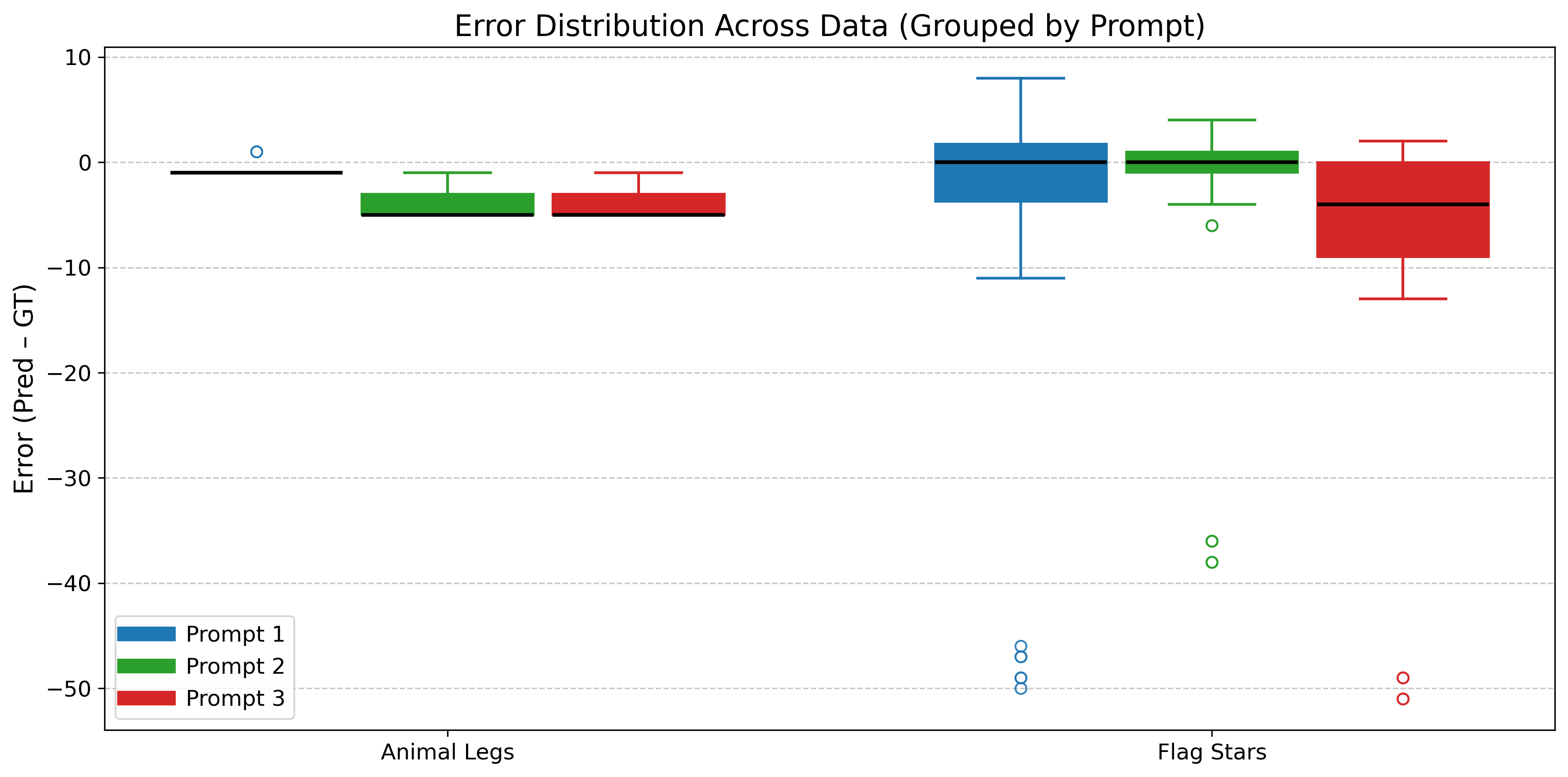}\\[-4pt]
                    \includegraphics[width=0.5\linewidth]{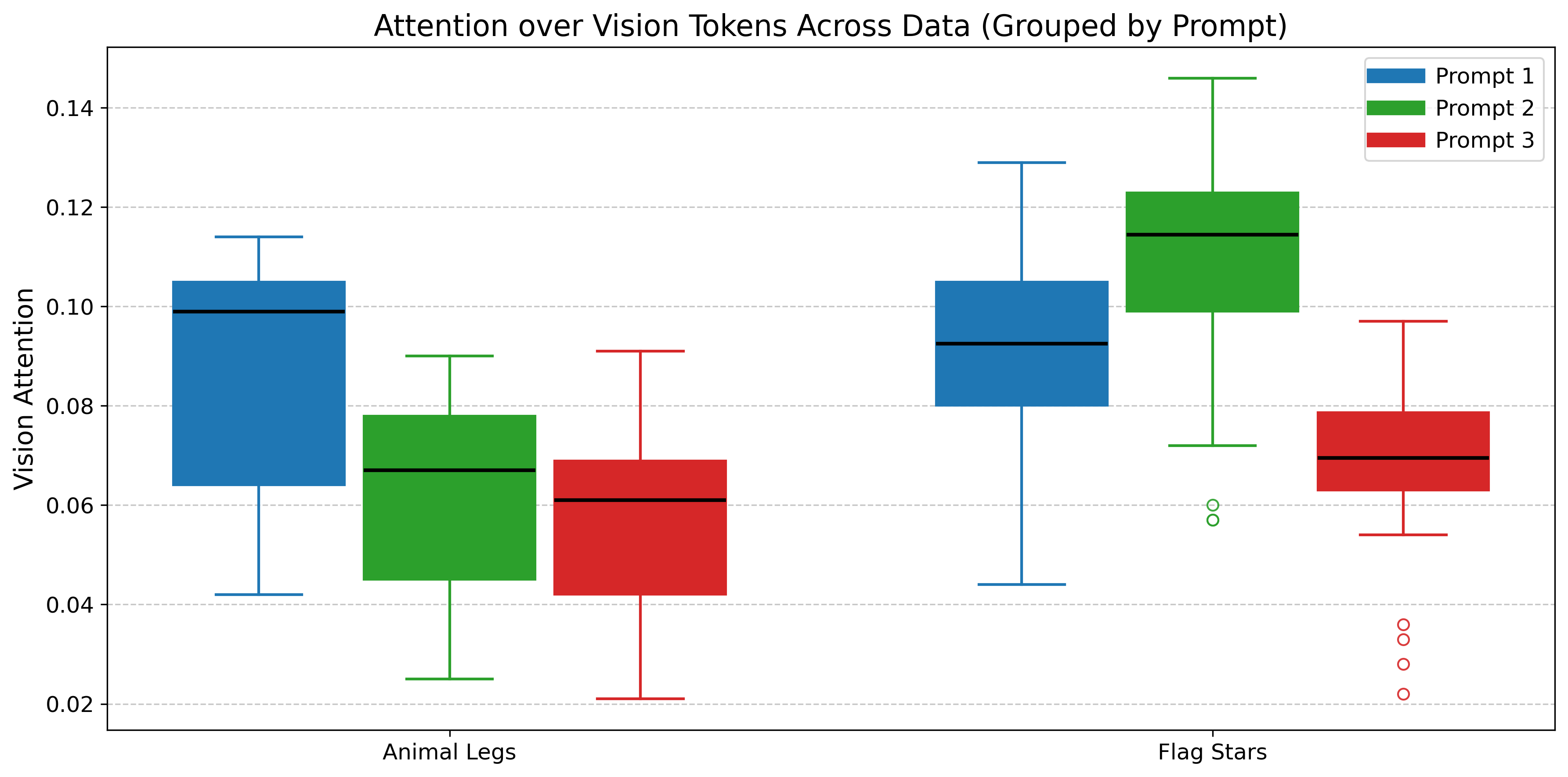}
                    & \includegraphics[width=0.5\linewidth]{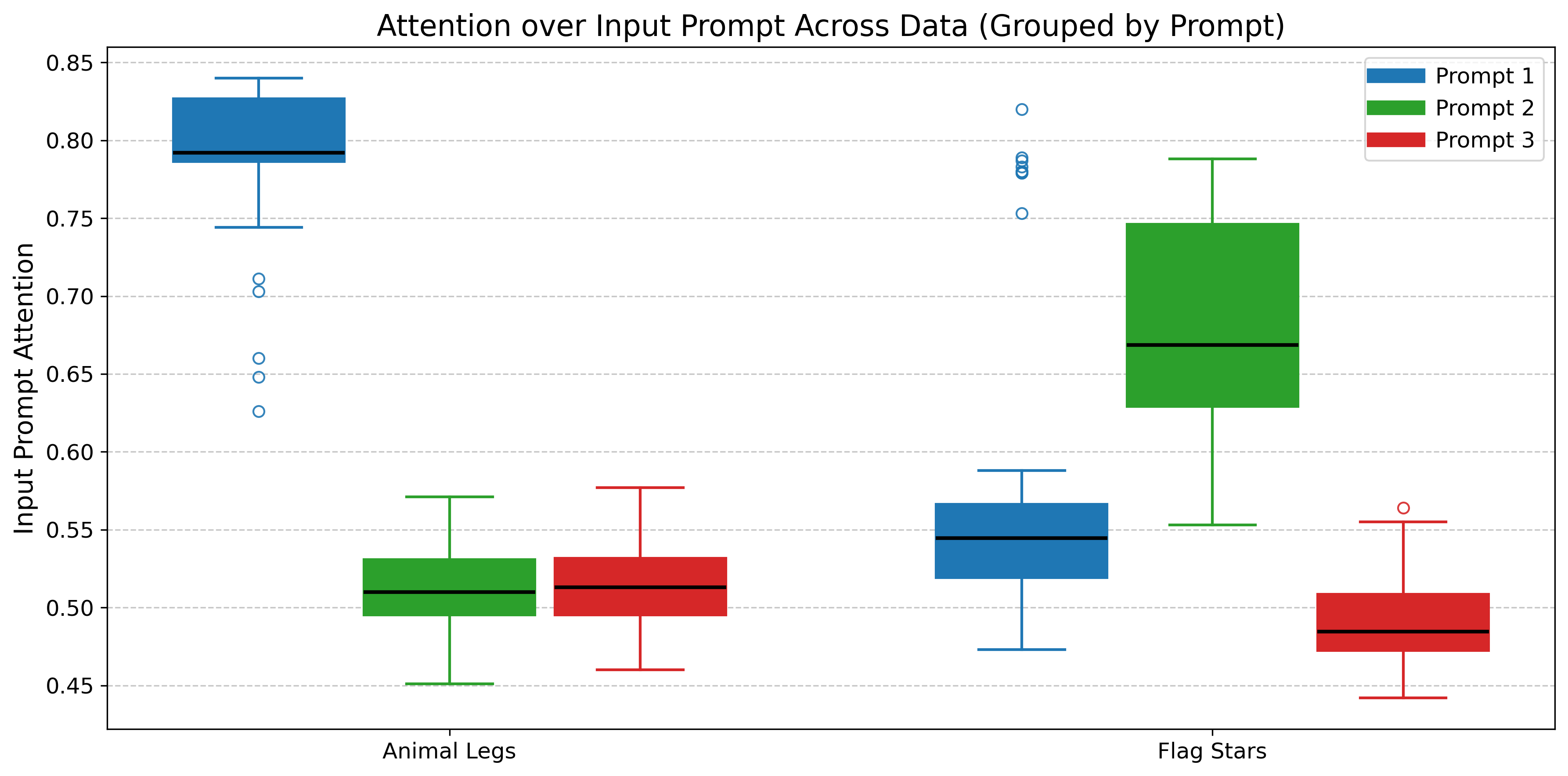}\\[-4pt]
                    \multicolumn{2}{c}{ \includegraphics[width=0.5\linewidth]{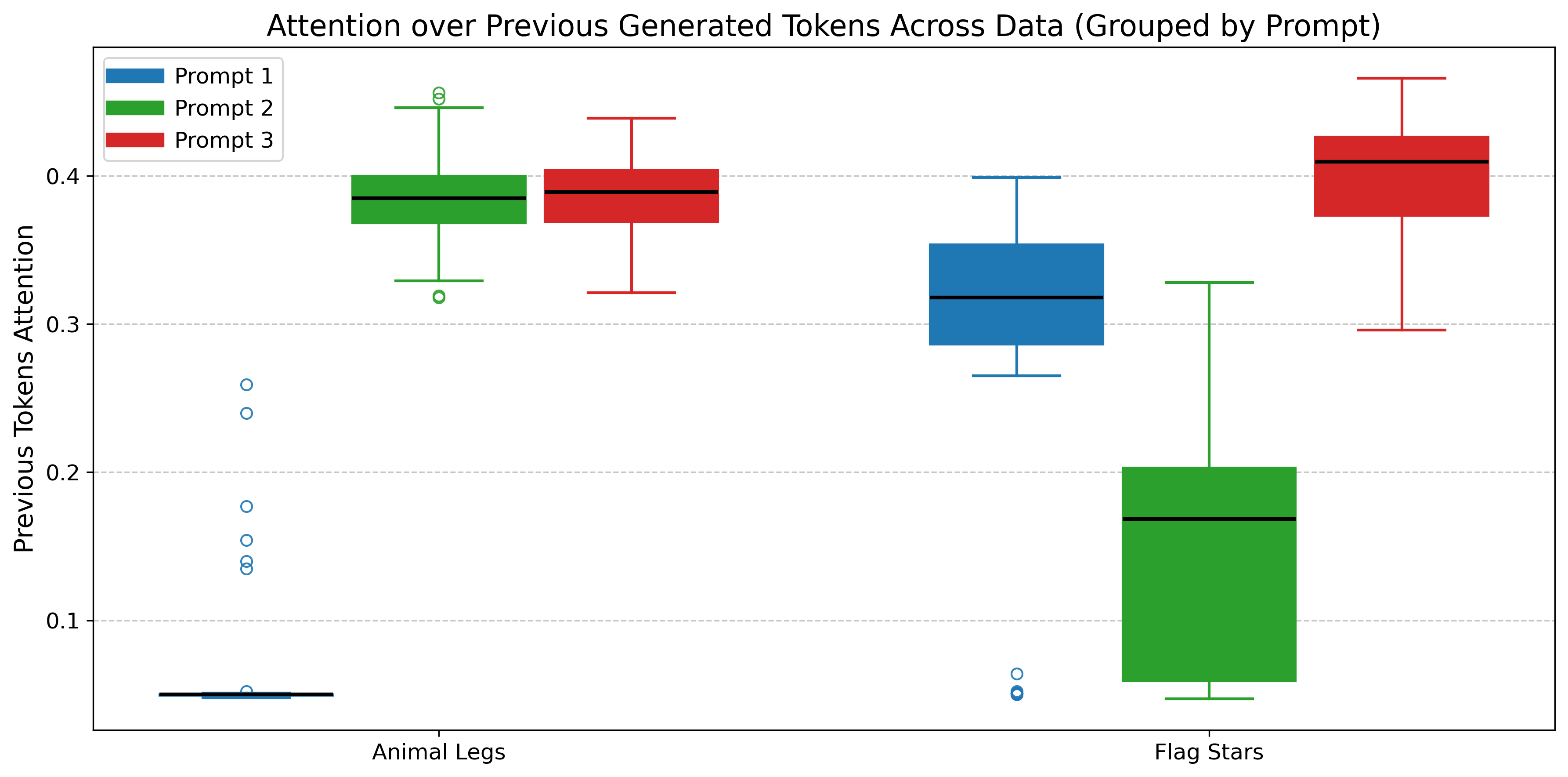}}
                    
                    \\
                \end{tabular}%
        \caption{Results on Qwen-2.5-VL-32B for data from `VLMs are Biased' paper from Vo et al. \cite{vlmsarebiased}}
        \label{fig:res_rwd_qwen32b}
\end{figure}
\begin{figure}[h]
            \centering
            
                \begin{tabular}{cc}
                     \includegraphics[width=0.5\linewidth]{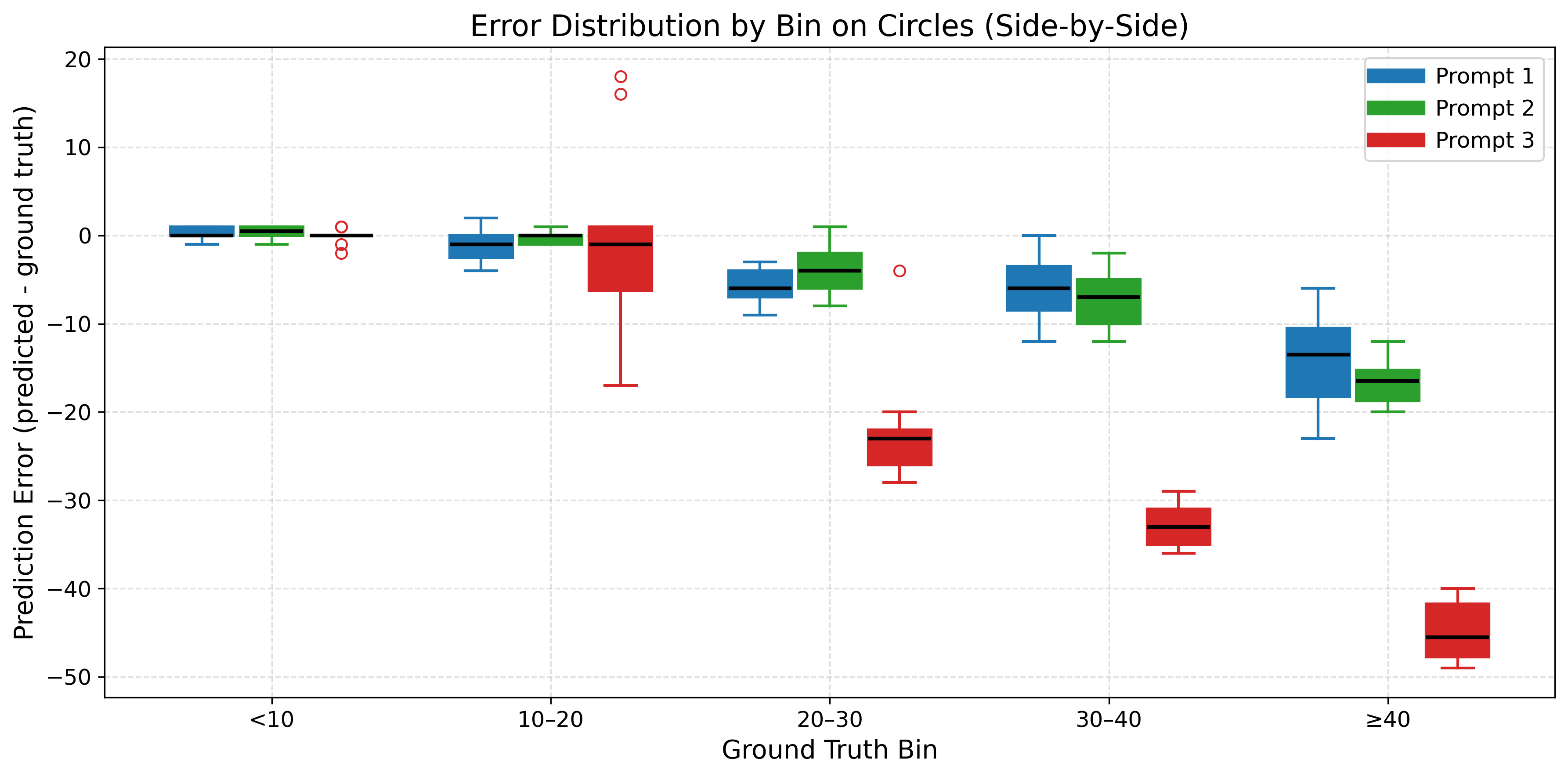}
                    & \includegraphics[width=0.5\linewidth]{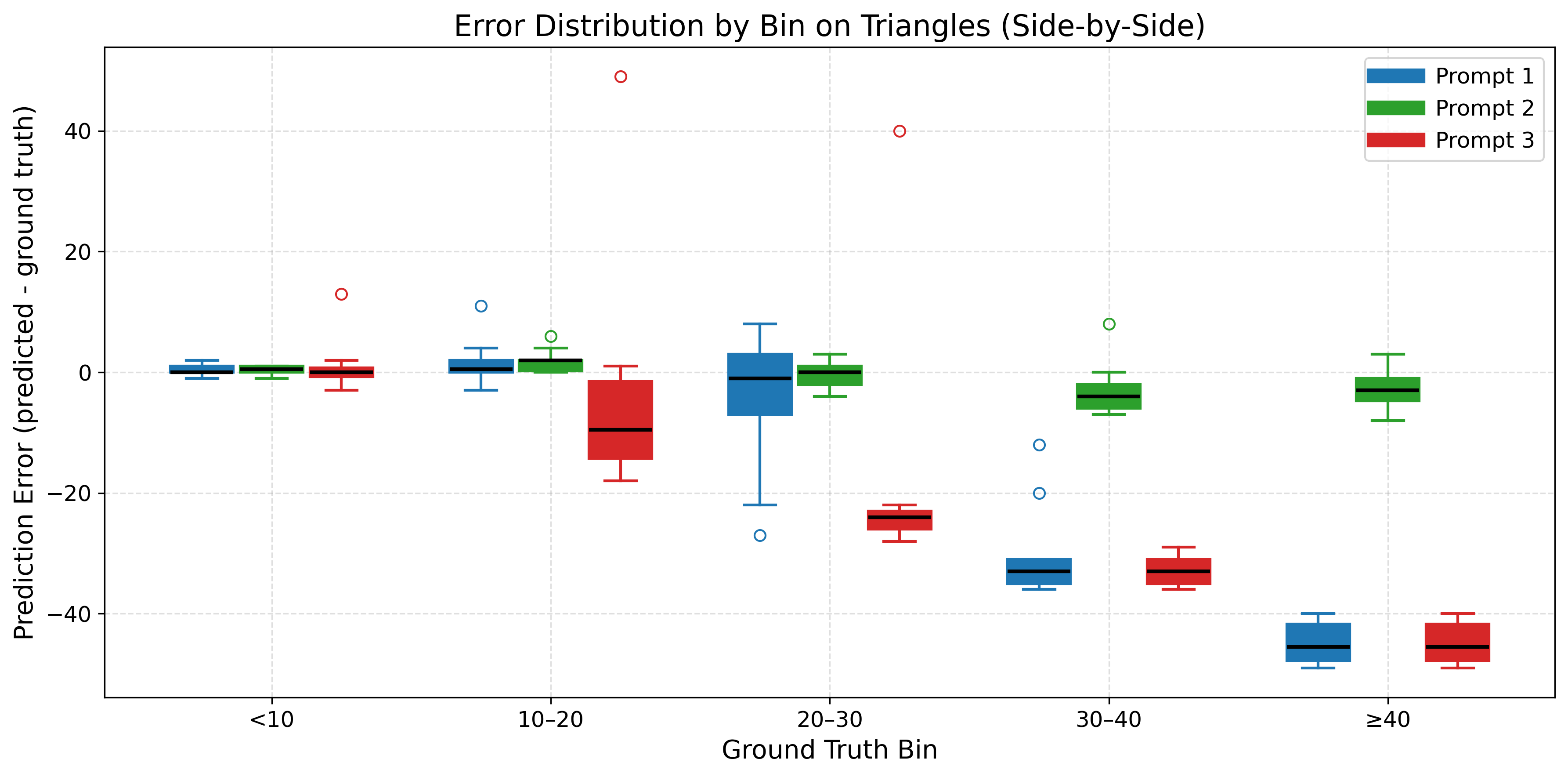}\\[-4pt]
                    \includegraphics[width=0.5\linewidth]{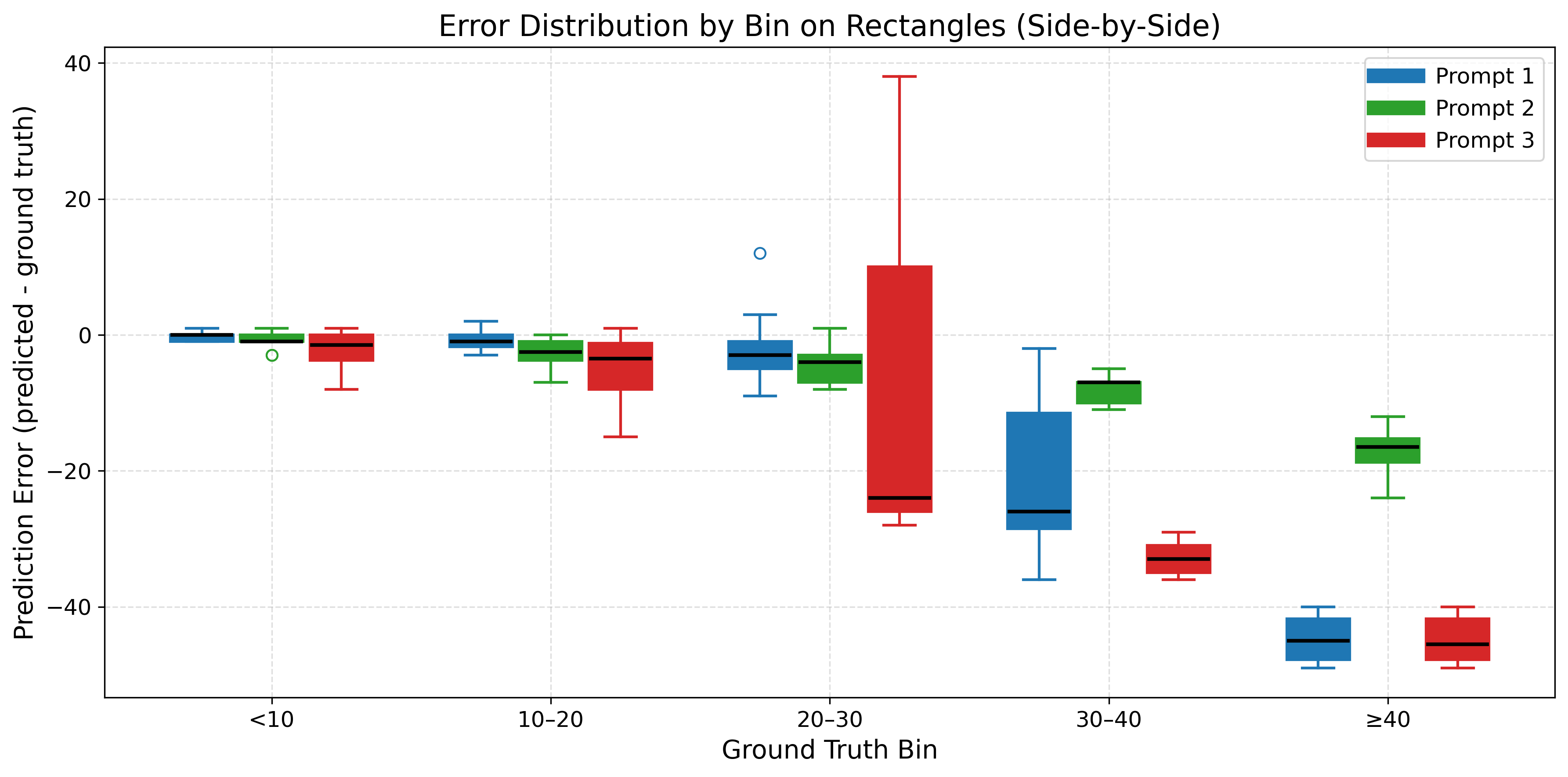}
                    & \includegraphics[width=0.5\linewidth]{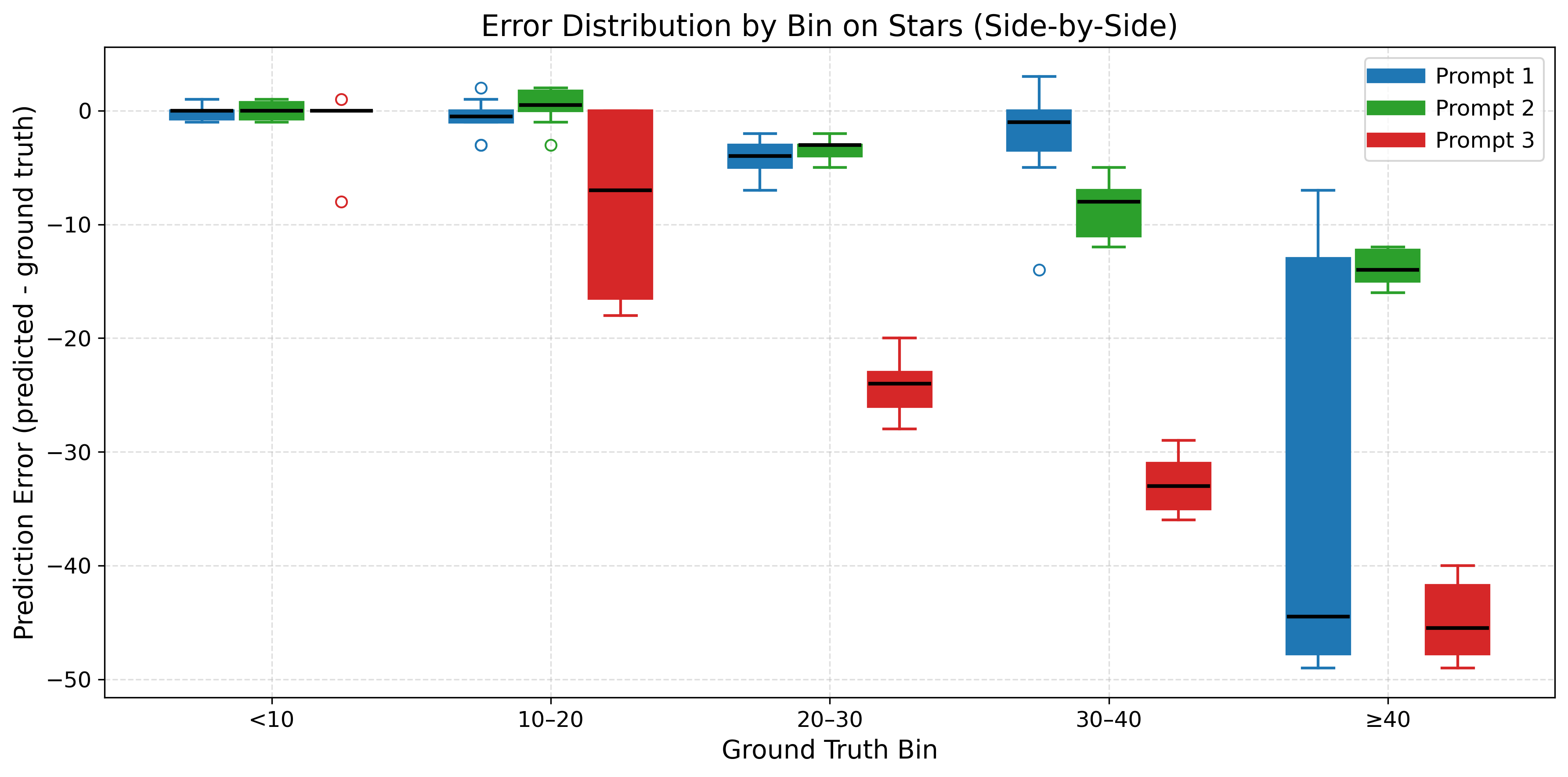}\\[-4pt]
                    \multicolumn{2}{c}{ \includegraphics[width=0.5\linewidth]{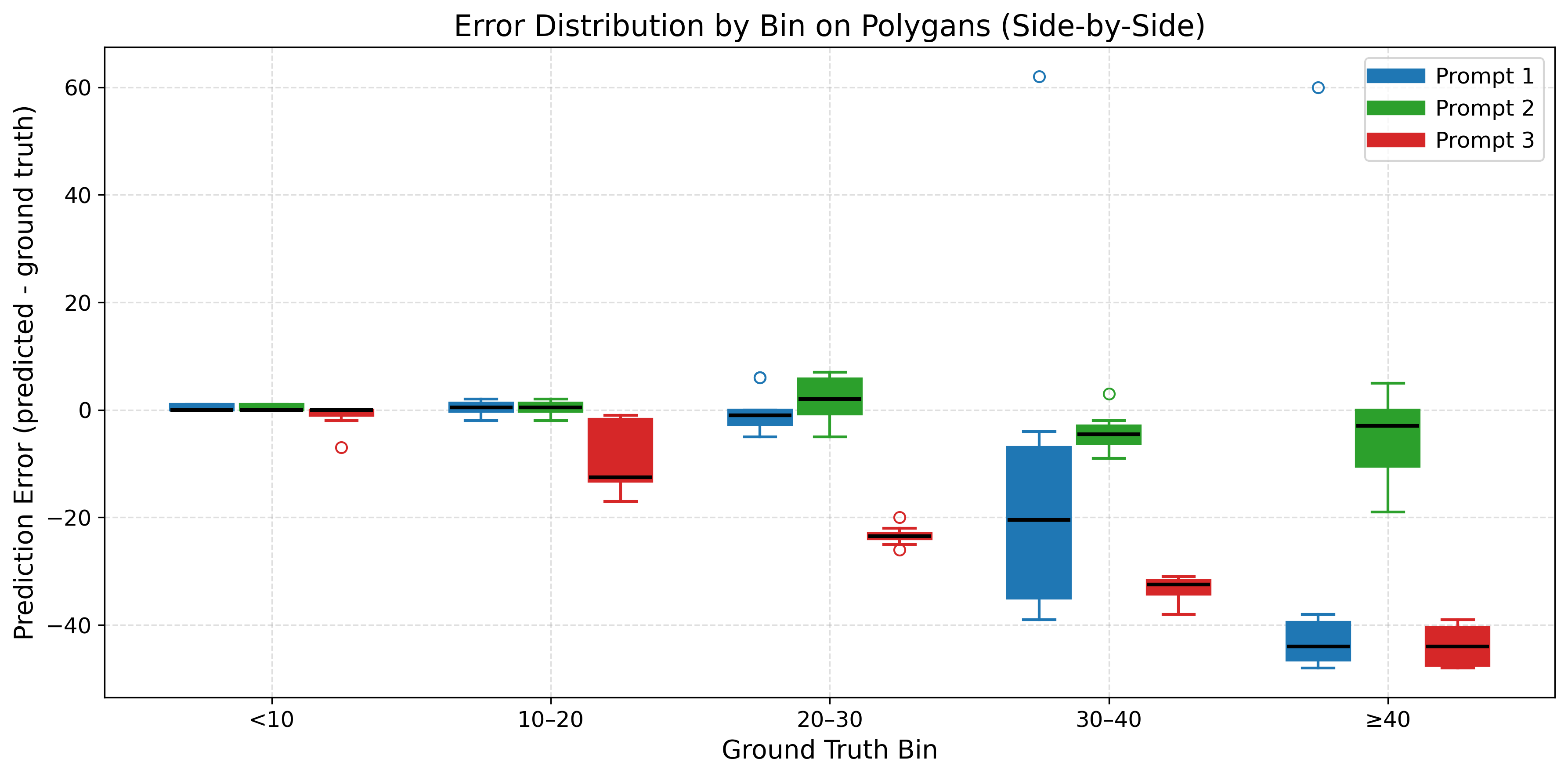}}
                    
                    \\
                \end{tabular}%
            \caption{Counting errors on Qwen-2.5-VL-7B}
            \label{fig:errors_qwen7b}
\end{figure}
\begin{figure}[h]
            \centering
            
                \begin{tabular}{cc}
                     \includegraphics[width=0.5\linewidth]{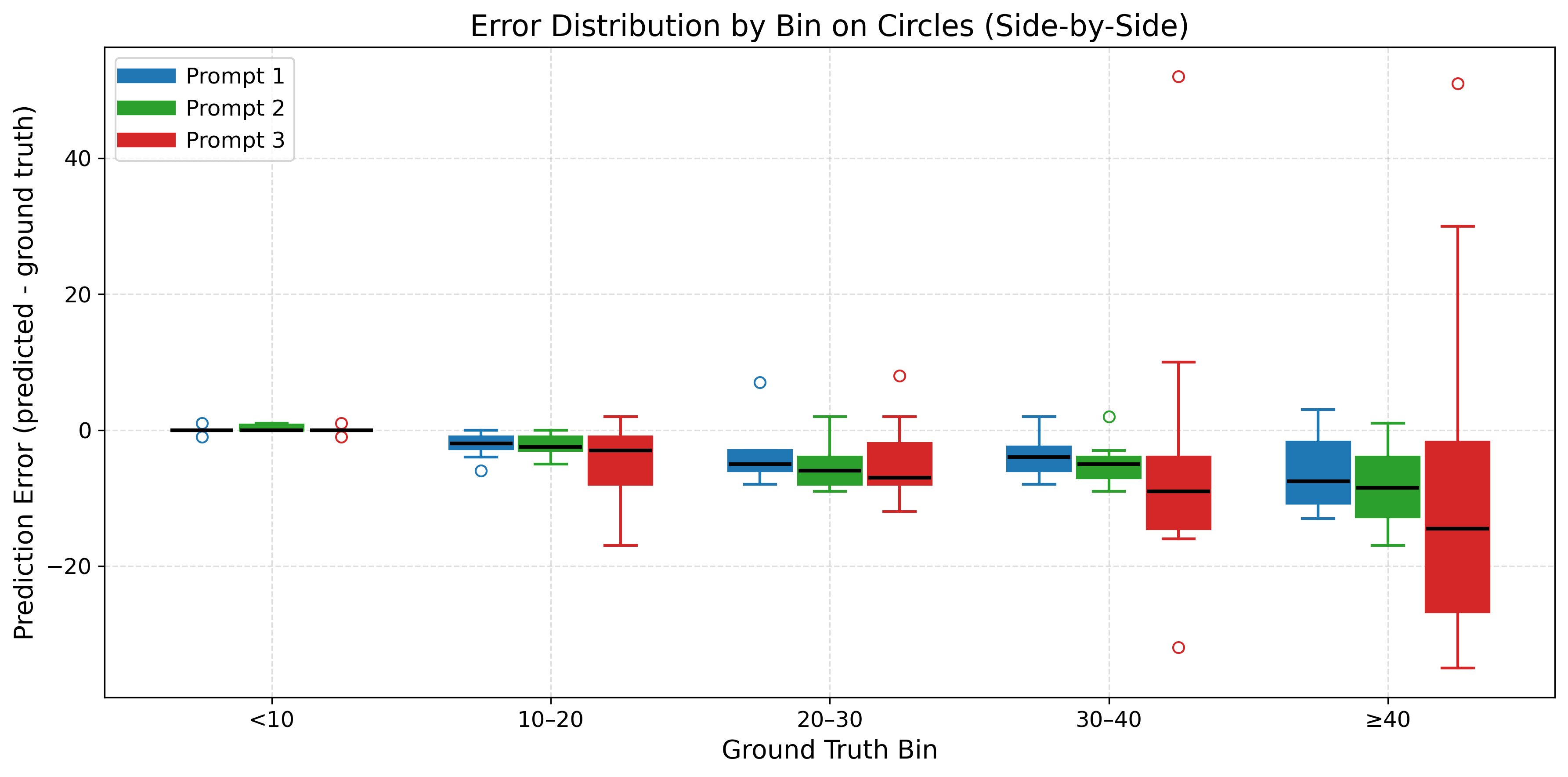}
                    & \includegraphics[width=0.5\linewidth]{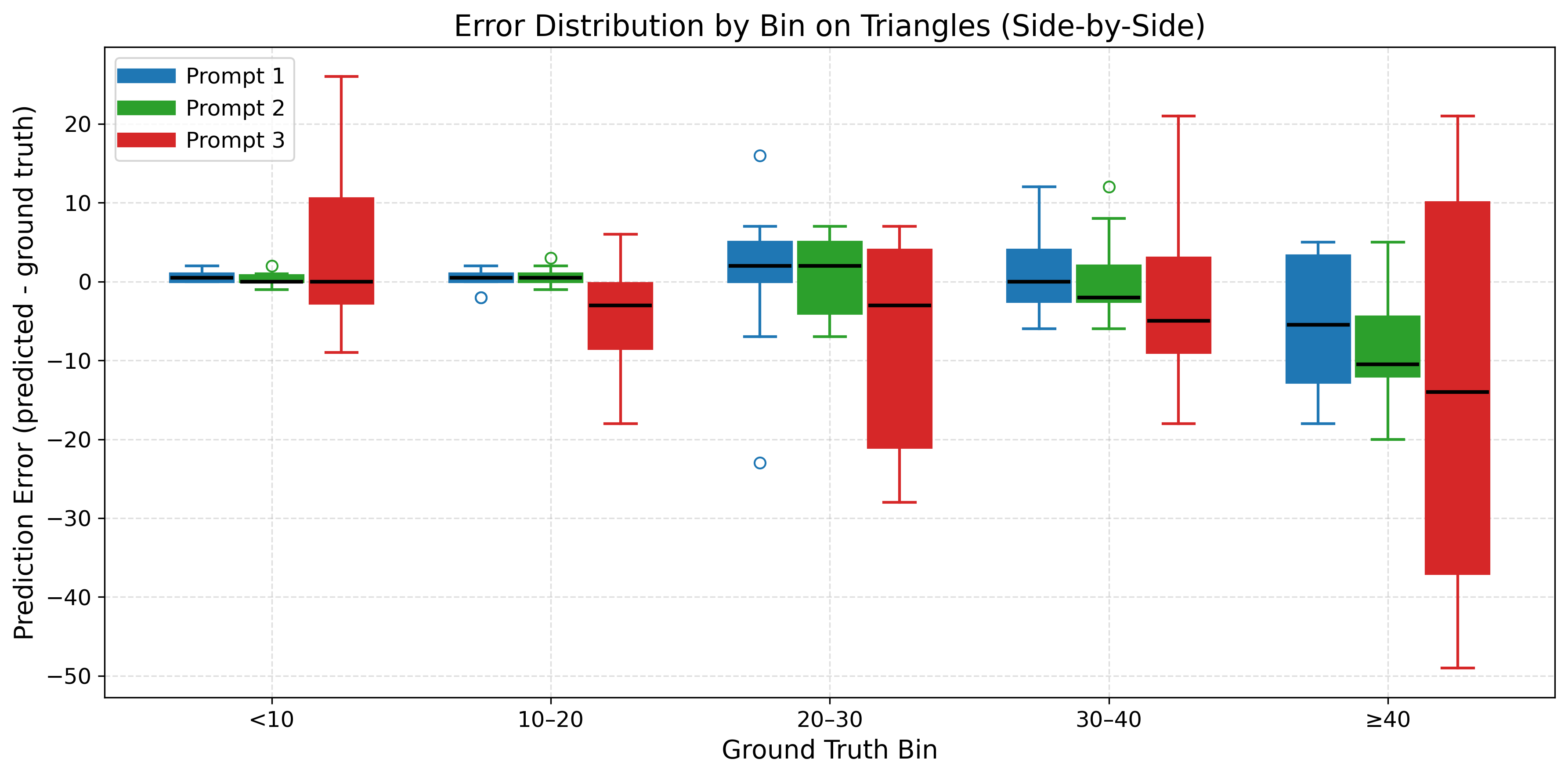}\\[-4pt]
                    \includegraphics[width=0.5\linewidth]{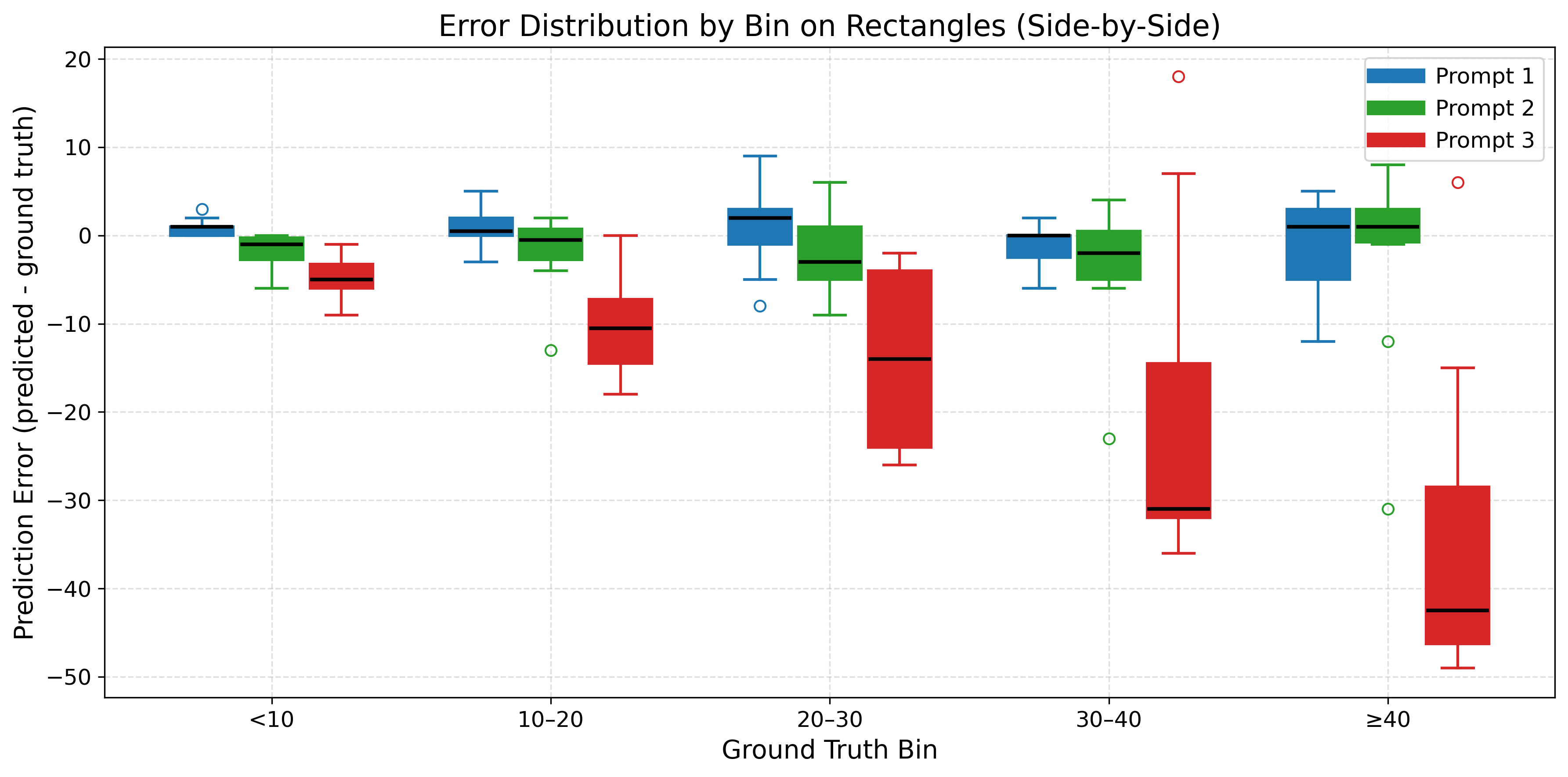}
                    & \includegraphics[width=0.5\linewidth]{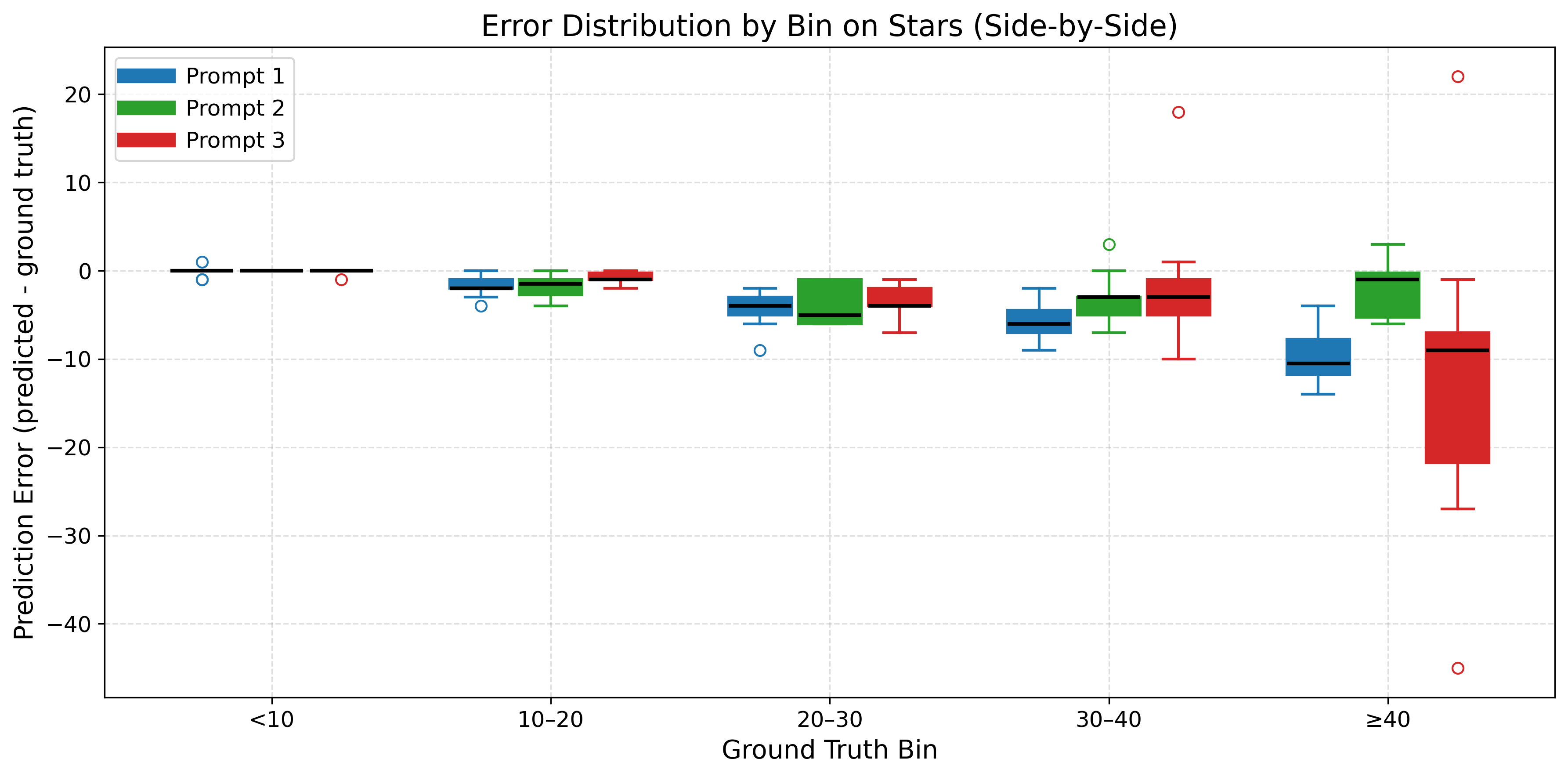}\\[-4pt]
                    \multicolumn{2}{c}{ \includegraphics[width=0.5\linewidth]{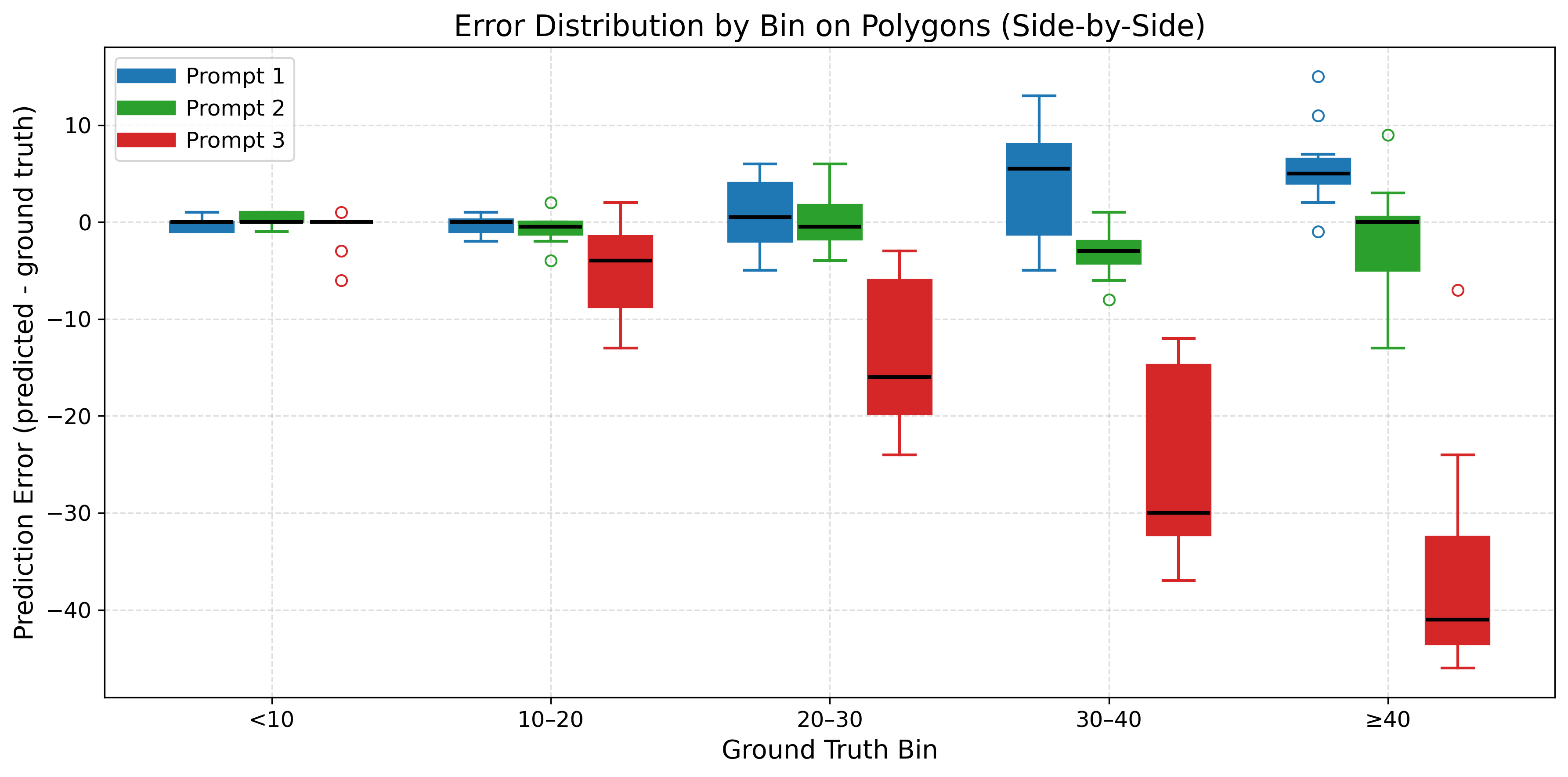}}
                    
                    \\
                \end{tabular}%
            \caption{Counting errors on Qwen-2.5-VL-32B}
            \label{fig:errors_qwen32b}
\end{figure}
\begin{figure}[h]
            \centering
            
                \begin{tabular}{cc}
                     \includegraphics[width=0.5\linewidth]{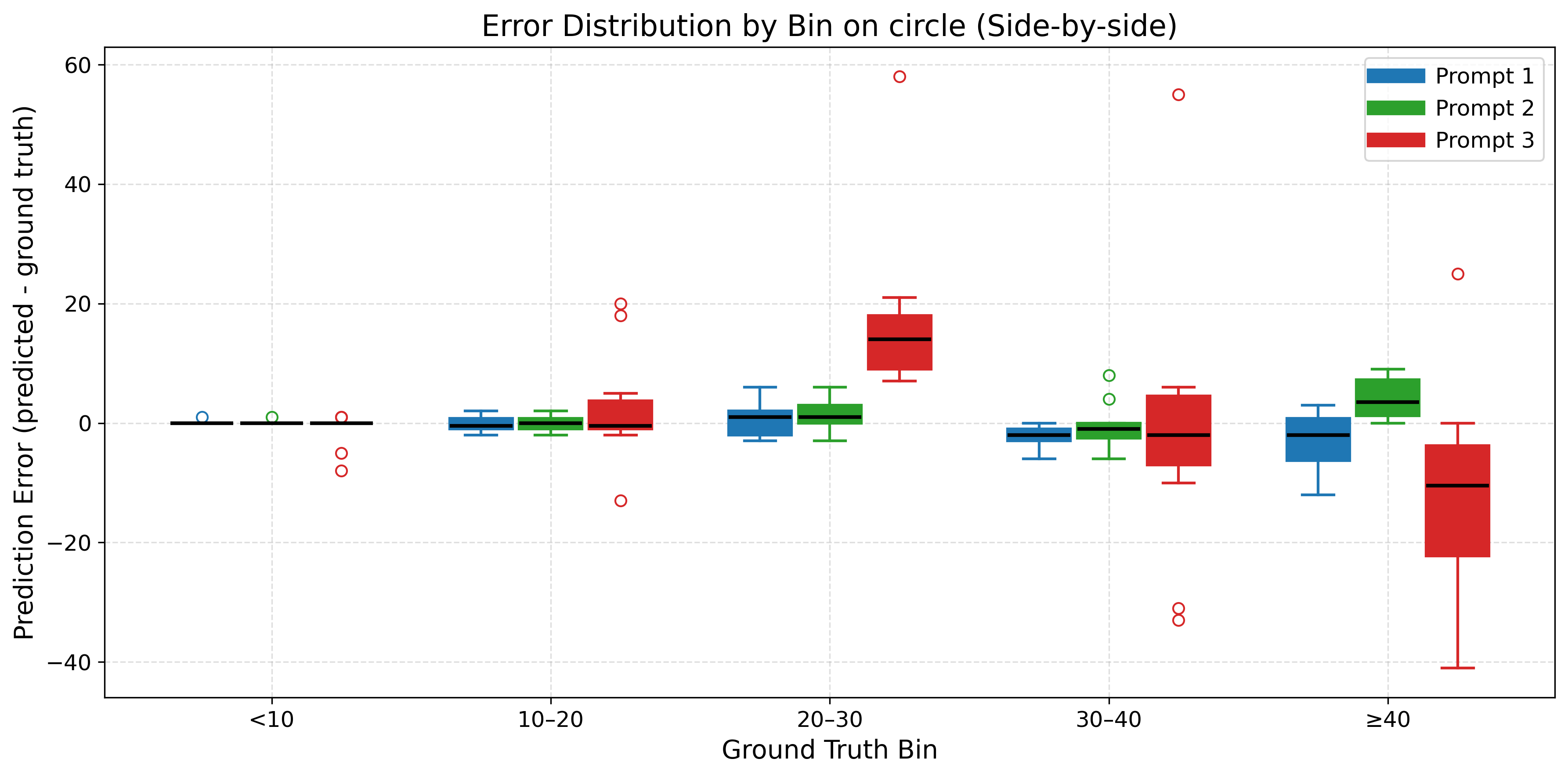}
                    & \includegraphics[width=0.5\linewidth]{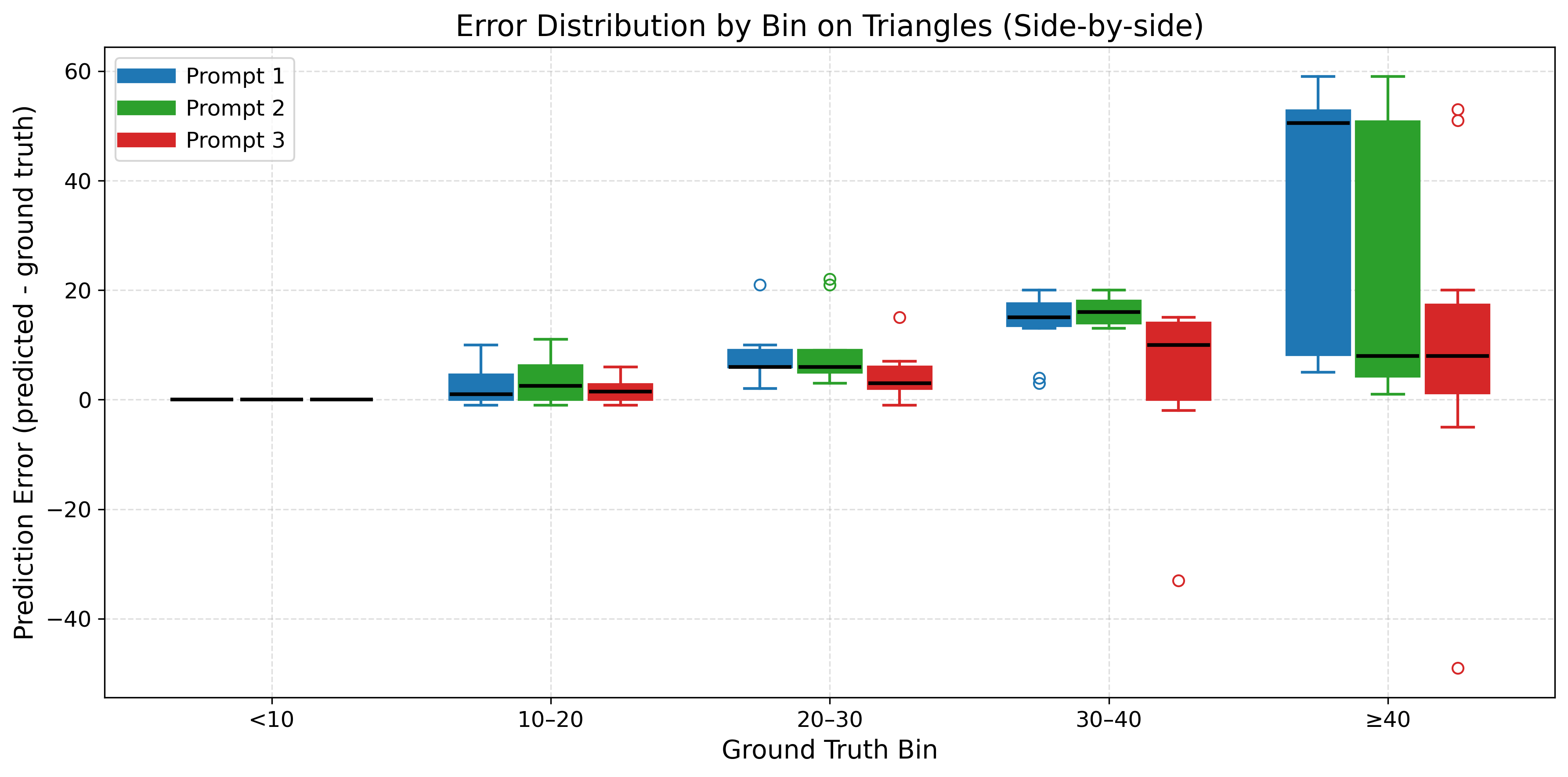}\\[-4pt]
                    \includegraphics[width=0.5\linewidth]{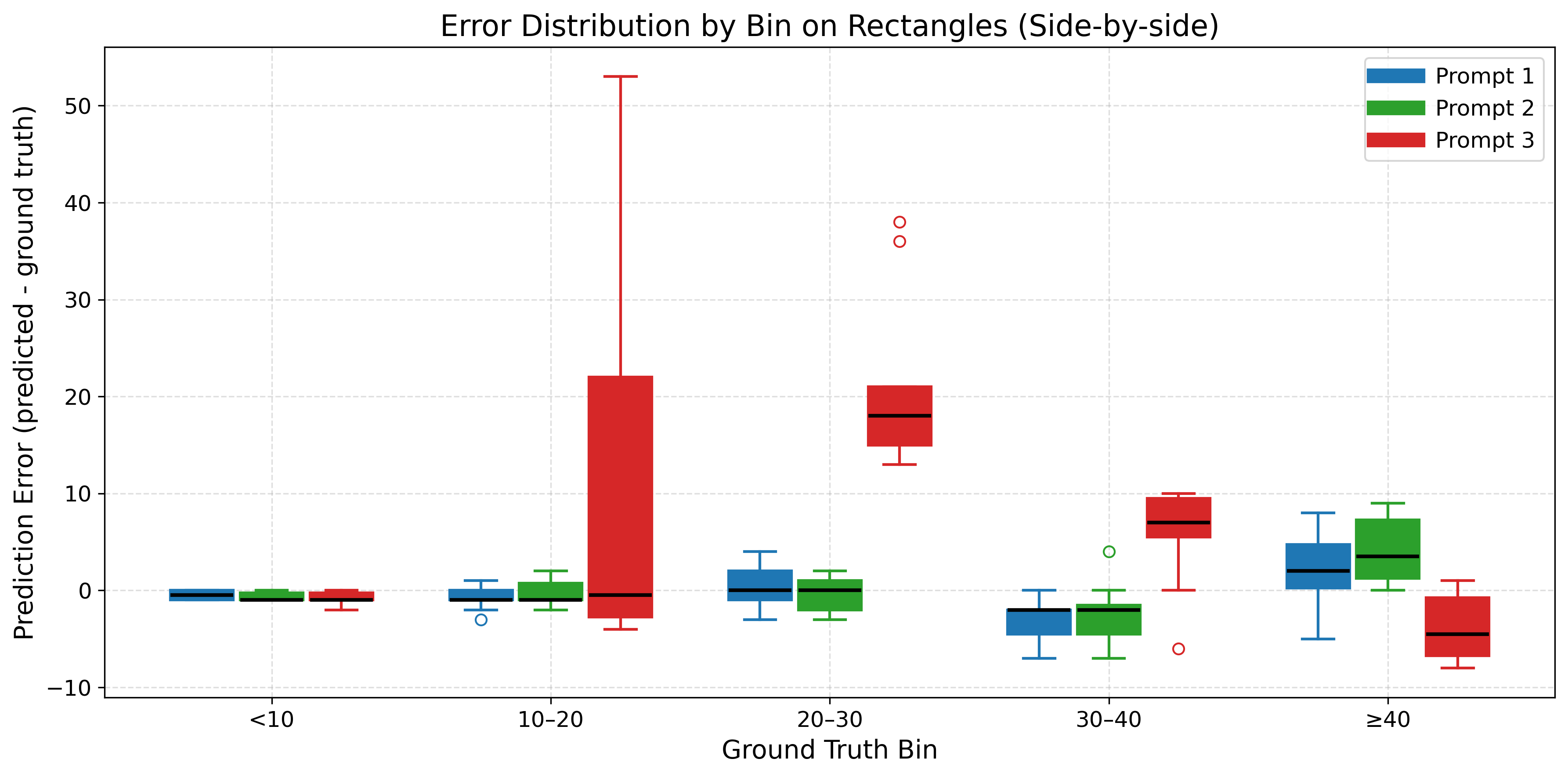}
                    & \includegraphics[width=0.5\linewidth]{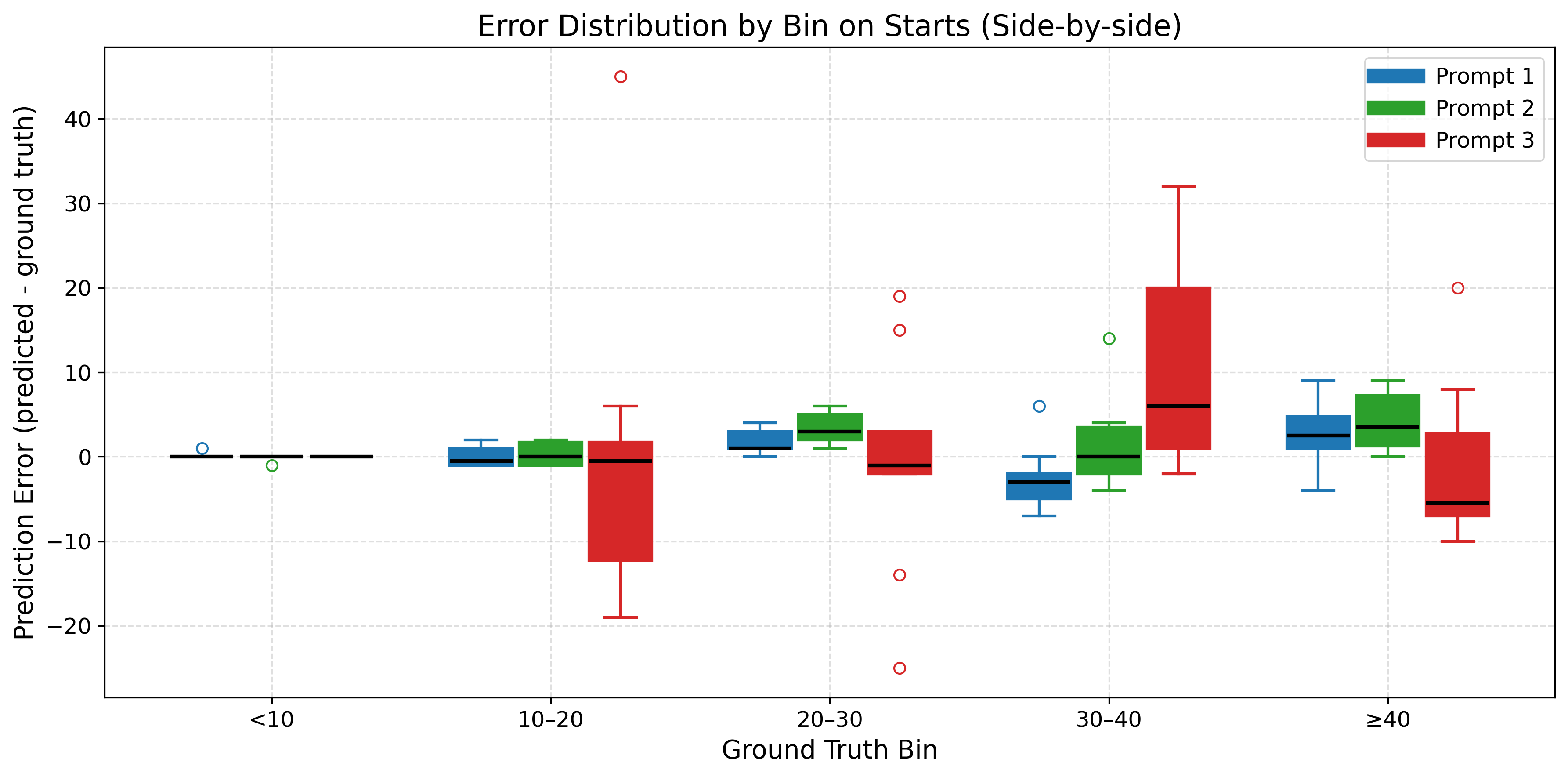}\\[-4pt]
                    \multicolumn{2}{c}{ \includegraphics[width=0.5\linewidth]{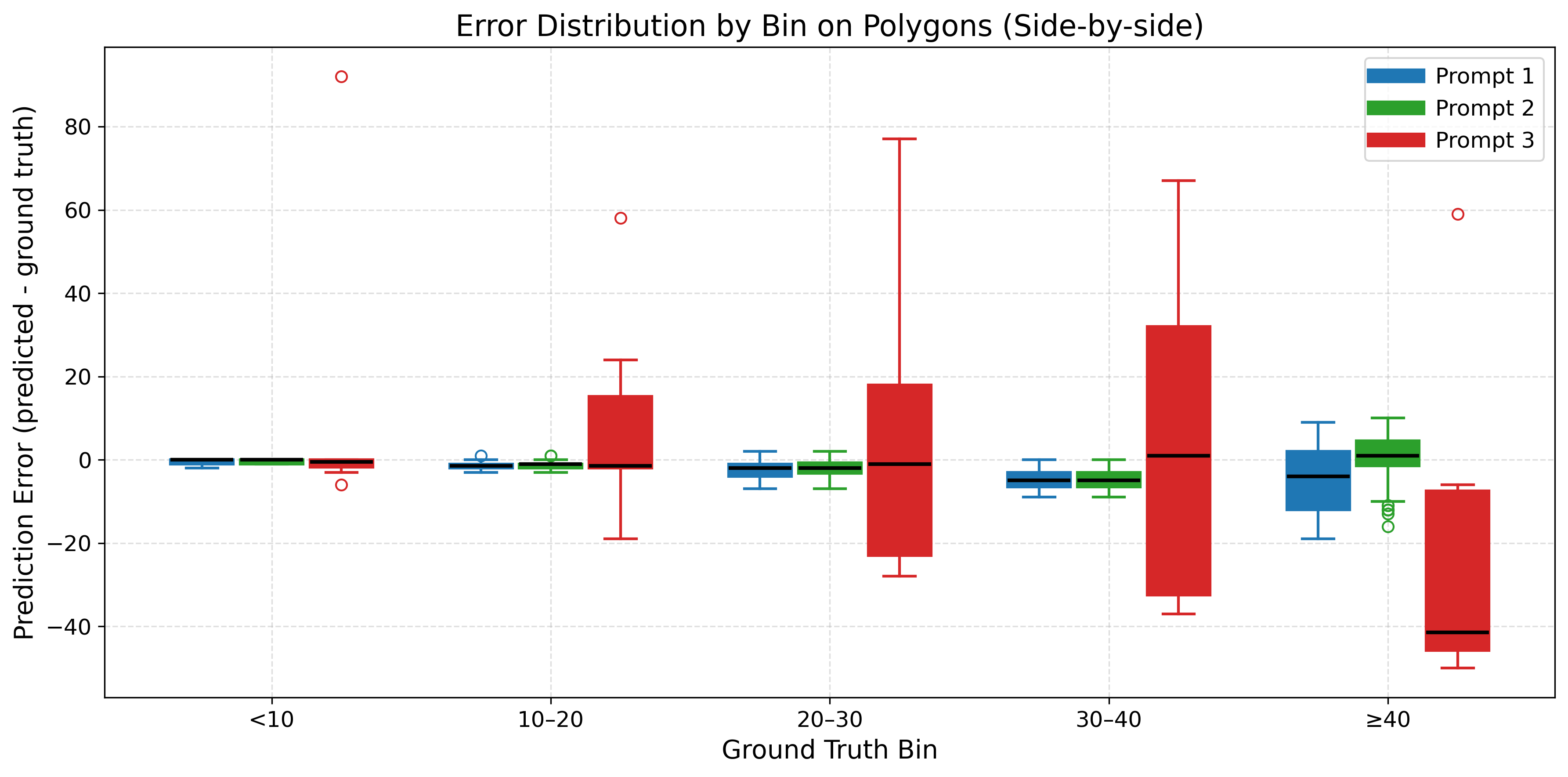}}
                    
                    \\
                \end{tabular}%
            \caption{Counting errors on Kimi-VL-A3B}
            \label{fig:errors_kimi}
\end{figure}

\begin{enumerate}
    \item For real-world data, the accuracy improves and the error decreases when prompts explicitly specify the target object (Prompts 2–3). In Flag Stars, specifying the target object and requiring structured output substantially increases accuracy (up to 0.4; Figures \ref{fig:res_rwd_qwen7b} and \ref{fig:res_rwd_qwen32b}). Under generic prompts, the model's attention is mostly on input text, while more specific prompts reduce this bias and shift attention toward previously generated tokens.
    \item We can see that when the counting prediction error (calculated using (Ground Truth Count - Predicted Count)) is higher, the model spends less of its total attention on input tokens; instead, it focuses more on generated output tokens. Refer Figures \ref{fig:result_qwen7b},\ref{fig:result_qwen32b} and \ref{fig:res_kimi}.

    \item Shape of the object in the image leads to very minor changes in the distribution of attention. We can see that the proportion of attention across the same prompt for different object shapes are within a very small interval. Refer Figures \ref{fig:result_qwen7b},\ref{fig:result_qwen32b} and \ref{fig:res_kimi}
    
    \item As the number of objects in the image increases, the counting errors increases. When the number of objects in the image is <10, the models perform relatively accurately, but counting performance becomes less accurate as we move towards the >40 bucket. Refer Figures \ref{fig:errors_qwen7b},\ref{fig:errors_qwen32b} and \ref{fig:errors_kimi}.

    \item Kimi-VL-A3B tends to overestimate counts, while Qwen2.5-VL tends to underestimate counts. Errors for Qwen 2.5-VL are centered mostly around negative values, meaning the model often underestimates compared to ground truth. While the opposite is true for Kimi-VL-A3B. Refer Figures \ref{fig:errors_qwen7b},\ref{fig:errors_qwen32b} and \ref{fig:errors_kimi}.
    
    \item The Qwen-7B model shows low and unstable accuracy with severe underestimation under detailed prompts, while the Qwen-32B model achieves higher and more stable accuracy with errors closer to zero, though both still struggle with fine-grained tasks like Animal Legs. Refer Figures \ref{fig:res_rwd_qwen7b} and \ref{fig:res_rwd_qwen32b}.

    \item Both Qwen2.5-VL and Kimi-VL-A3B models assign relatively little attention to vision tokens. 
    \item We replicate the results for data from Vo et al. \cite{vlmsarebiased}, when using the same prompts and data as them. Refer Figures \ref{fig:res_rwd_qwen7b} and \ref{fig:res_rwd_qwen32b}.

\end{enumerate}

\subsection{Conclusion and Future Work}

Our study shows that VLMs are highly sensitive to both image characteristics and prompt specificity. Importantly, we find that bias from prior knowledge can be partially addressed by specifying prompts more precisely, which shifts attention and improves model performance. Future work includes probing more visual dimensions (e.g., color, background, occlusion), automating prompt design, combining attention with gradient-based interpretability, and extending evaluation beyond counting to reasoning tasks.

\end{document}